\documentclass[10pt,letterpaper]{article}
\usepackage[shownumpages,citingstyle=authoryear,bibliostyle=plainnat,bibfile=refs,boxrule=0.7pt]{brainlab}

\usepackage{amsmath,amsfonts,bm}

\def\eqref#1{equation~\ref{#1}}
\def\1{\bm{1}}

\DeclareMathAlphabet{\mathsfit}{\encodingdefault}{\sfdefault}{m}{sl}
\SetMathAlphabet{\mathsfit}{bold}{\encodingdefault}{\sfdefault}{bx}{n}

\usepackage{hyperref}
\usepackage{url}

\usepackage{amsmath}          % расширенная математика (align, cases и т.д.)
\usepackage{amssymb}          % дополнительные математические символы
\usepackage{mathtools}        % надстройка над amsmath (доп. окружения, умные скобки)
\usepackage{graphicx}         % вставка изображений
\usepackage{booktabs}         % профессиональные таблицы (\toprule, \midrule, \bottomrule)
\usepackage{array}            % расширенные возможности таблиц
\usepackage{colortbl}         % \rowcolor, \cellcolor — подсветка строк/ячеек таблиц
\usepackage{makecell}         % \makecell — перенос строк внутри ячейки таблицы
\usepackage{enumitem}         % гибкая настройка списков (itemize/enumerate)
\usepackage{subcaption}       % подрисунки (subfigure/subtable)
\usepackage{multirow}         % объединение ячеек таблицы по строкам
\usepackage{tabularx}         % таблицы с автоматической шириной столбцов
\usepackage{float}
\usepackage{wrapfig}          % обтекание текстом рисунков/таблиц
\usepackage{adjustbox}        % масштабирование таблиц/рисунков под ширину колонки
\usepackage{xcolor}           % цвета (\definecolor, \textcolor)
\usepackage{cleveref}         % умные ссылки \cref{} вместо \ref{} — подключать последним, после hyperref
\graphicspath{{figures_dense/}{figures_moe/}{figures/}}        % изображения ищутся в подпапке figs/

\tcbuselibrary{most}  % compact boxes for the main research question
\newtcolorbox[auto counter]{researchquestion}{%
  enhanced,
  colback=gray!4,
  colframe=gray!55,
  boxrule=0.5pt,
  arc=1mm,
  left=6pt,
  right=6pt,
  top=5pt,
  bottom=5pt,
  before skip=8pt,
  after skip=8pt,
  fonttitle=\bfseries,
  title=Research question~\thetcbcounter}

\makeatletter
\renewcommand{\subsection}{%
  \@startsection{subsection}{2}{\z@}%
  {-1.6mm}{0.7mm}{\normalfont\normalsize\bfseries}}
\renewcommand{\paragraph}{%
  \@startsection{paragraph}{4}{\z@}%
  {-1.6mm}{-2mm}{\normalfont\normalsize\bfseries}}
\makeatother

\setbrainmeta{
 title={Not All Drift Is a Mechanism: \\ Tracing Prompt Adaptation in LLMs},
 authors={{\normalfont\mdseries\normalsize
\setlength{\tabcolsep}{7pt}
\begin{tabular}{@{}lll@{}}
\textbf{Mikhail Aleksandrov\equalcontrib} & \textbf{Nail Bashirov\equalcontrib} & 
\textbf{Maksim Ivanov}\\[0.7em]
\textbf{Andrew Tsukanov} & \textbf{Timofey Karyagin} & \textbf{Alexander Miasnikov}\\[0.7em]
\textbf{Gleb Molodtsov} & \textbf{Valentin Samokhin} & \textbf{Aleksandr Beznosikov}
\end{tabular}
}},
 affiliations={%
  \equalcontribfalse
  {\small
  BRAIn Lab \\
  \textsuperscript{*}Equal contribution}\vspace{1mm}
},
  code = {\url{https://github.com/brain-lab-research/prompt-adaptation}},
 abstract={Prompt-based adaptation offers several ways to specialize a frozen language model, ranging from natural-language prompt optimization to learned continuous prompts. These methods are usually compared through downstream performance, although similar outputs need not arise from similar internal computations. We study how discrete prompt optimization, prompt tuning, and prefix tuning alter the processing of dense and mixture-of-experts language models. Our analysis shows that the location of the adaptation predicts how task information is carried to the output: optimized text primarily changes response construction, input-level prompts write their influence into content representations, and layer-wise prefixes maintain direct control through attention. In MoE models, these differences further induce distinct expert-routing patterns. At the same time, common representation-level measurements respond strongly to generic context changes and do not reliably distinguish useful adaptations from matched placebos. We argue that understanding prompt adaptation requires separating representation drift from the part of that drift that is aligned with, and causally used by, the model's task computation. This perspective provides practical guidance for selecting, evaluating, and interpreting prompt adaptation methods.}
}
\begin{document}

\maketitlebox

\section{Introduction}

Large language models (LLMs) are increasingly adapted as frozen backbones, allowing one model to support many tasks without storing or serving a separate set of weights for each application. Context becomes the adaptation interface: a task can be specified through a natural-language instruction, learned continuous vectors at the input, or learned states inserted into the attention computation~\citep{agrawal2026gepa,lester2021prompt,li2021prefix}.

The most direct approach searches for a better textual prompt. Early methods optimized individual tokens, while recent systems use language models to propose, critique, and refine complete instructions~\citep{shin2020autoprompt,pryzant2023automatic}. A complementary line replaces text with trainable continuous states. Prompt tuning learns input embeddings, whereas prefix tuning exposes learned keys and values to every Transformer layer~\citep{lester2021prompt,li2021prefix}. These methods optimize the context seen by the same frozen model, but provide different access to its computation.

In practice, these methods are usually compared by downstream score. This is insufficient for understanding what the adaptation changed. A gain may arise from a better task decision, more reliable adherence to an output protocol, or a different route through which an existing decision reaches the language-model head. In a mixture-of-experts (MoE) model, the same contextual change can additionally alter which expert parameters process each token~\citep{olson2025semanticrouting}. Similar outputs therefore need not imply similar internal computations.

Research on prompt optimization has made substantial progress in improving task performance, while mechanistic studies of contextual learning have identified task vectors and prompt-induced geometric changes~\citep{agrawal2026gepa,todd2024function,kirsanov2025geometry}. These directions have largely developed separately. Performance-oriented studies rarely locate the computation that mediates the gain, while representation analyses often show that a prompt changes hidden states without testing whether the measured change is sufficient to alter the output~\citep{hewitt2019probes,zhang2024patching}.

We study prompt adaptation from this functional perspective. We compare discrete prompt optimization, prompt tuning, and prefix tuning under a common frozen-model protocol, follow task information through a dense model, and test observational signatures against matched placebos and causal interventions. We then examine sparse models, where a hidden-state change can become a discrete routing decision. Our results support a common account: the insertion site constrains how an adaptation can influence the network, while common metrics often capture broad representation drift rather than the smaller state change sufficient to mediate the adapted output.

\section{Related Work}

\paragraph{Continuous prompts.}
Continuous prompting adapts a frozen model by optimizing trainable states rather than model parameters. Prefix tuning introduced task-specific keys and values that remain available throughout the attention stack~\citep{li2021prefix}, while prompt tuning showed that embeddings inserted only at the input become increasingly competitive as model scale grows~\citep{lester2021prompt}. Deep variants such as P-Tuning v2 extend these states across layers and tasks~\citep{liu2022ptuningv2}. These methods therefore differ not only in parameter count, but also in the depth at which they can influence computation. Theoretical analyses characterize both the expressive power of prompt-mediated attention and the restrictions imposed by a frozen backbone~\citep{oymak2023attention,petrov2023prompting}. Their learned states remain difficult to interpret, since nearest-token projections can be unfaithful and human-readable constraints may reduce task performance~\citep{patel2025interpretable}.

\paragraph{Discrete prompt evolution.}
A parallel line retains the natural-language interface and searches for better instructions. Prompt sensitivity first motivated gradient-guided trigger selection in AutoPrompt~\citep{shin2020autoprompt} and reward-based token generation in RLPrompt~\citep{deng2022rlprompt}. With stronger instruction-following models, optimization moved from isolated tokens to complete prompts: APE samples and selects candidate instructions~\citep{zhou2023ape}, ProTeGi combines textual gradients with beam search~\citep{pryzant2023automatic}, and OPRO conditions an optimizer on previous solutions and scores~\citep{yang2024opro}. Later methods jointly optimize instructions, demonstrations, or multistage programs~\citep{opsahlong2024mipro}. GEPA adds natural-language reflection and evolutionary selection~\citep{agrawal2026gepa}, while SPRIG studies iterative system-prompt optimization and transfer~\citep{zhang2026sprig}. This progression makes discrete adaptation accessible without gradients through the task model, but most evaluations still describe the resulting behavior rather than the computation that produced it.

\paragraph{Mechanistic analysis.}
Mechanistic studies of contextual learning first asked which computations prompts elicit. Induction heads identify circuits that copy contextual patterns~\citep{olsson2022induction}, simplified Transformers can implement optimization-like updates during the forward pass~\citep{vonoswald2023gradient}, and task or function vectors can transfer behavior under intervention~\citep{hendel2023taskvectors,todd2024function}. This work motivated observational diagnostics of prompt-induced computation, including representation geometry~\citep{kirsanov2025geometry}, task descriptions recovered through Patchscopes~\citep{ghandeharioun2024patchscopes,ramati2025inspect}, and hidden-state shifts after system-prompt optimization~\citep{zhang2026sprig}. Such measurements establish that context changes the network, but not that the measured change produces the output. Linear probes and the tuned lens localize decodable information~\citep{hewitt2019probes,belrose2023tunedlens}; activation steering and patching test causal sufficiency~\citep{turner2023activation,zhang2024patching}; and sparse autoencoders provide a feature basis in which these interventions can be expressed~\citep{cunningham2023sparse,gao2024scaling}. Our comparison brings these stages together for three adaptation mechanisms under the same task and backbone, using matched placebos to test observational measurements and causal transfer to identify state-level mediators.

\paragraph{MoE routing.}
This distinction becomes more explicit in mixture-of-experts models, where a contextual change can alter which parameters process a token. MoE Transformers route each token through a small expert subset, increasing total capacity at approximately fixed active computation~\citep{fedus2022switch}. Load-balancing objectives limit concentrated traffic and unused experts~\citep{wang2024lossfree, tang2025pangu, molodtsov2026hierarchical}, while recent analyses show that routing reflects both token identity and contextual semantics~\citep{olson2025semanticrouting}. Existing adaptation work mainly modifies weights, routers, or routed adapter modules~\citep{liu2025routedpeft}. The response of a frozen router to discrete, input-level, and layerwise prompt adaptation remains underexplored.

% ===== Inlined from ICLR/dense_section_1 =====
\newcommand{\TODO}[1]{\textcolor{red}{[TODO: #1]}}

\paragraph{Contributions.} Our contributions can be summarized as follows:

\begin{itemize}[nosep]
  \item \textbf{A mechanistic explanation of prompt adaptation.}
  We go beyond performance comparisons to explain how GEPA, prompt tuning, and prefix tuning change a frozen model's computation. We trace their influence through attention, hidden states, and expert routing, using controls and interventions to separate useful adaptation from generic context effects.

  \item \textbf{How adaptation acts through attention in dense models.}
  We study dense models to examine how each method changes attention and carries its influence into content representations. Attention masking and state interventions reveal how these changes affect output, while matched placebos show why representation drift alone is insufficient evidence of a useful mechanism.

  \item \textbf{How adaptation changes computational paths in MoE models.}
  We examine the additional lever provided by routing: the same token can pass through different expert networks after adaptation. We separate changes in expert selection from changes in serving load, then test expert necessity through pruning. These interventions expose adapter-specific dependencies on rarely used experts that traffic statistics alone miss.
\end{itemize}

\section{Dense Models: From Adaptation Interface to Functional State Change}
\label{sec:dense}

GEPA, prompt tuning, and prefix tuning all optimize context while keeping the
language model frozen, but they expose different parts of its computation to
that optimization. GEPA rewrites natural-language tokens, prompt tuning learns
embeddings inserted only at the input, and prefix tuning supplies learned keys
and values to every attention block
\citep{agrawal2026gepa,lester2021prompt,li2021prefix}. These insertion sites
define distinct routes by which an adaptation can remain available during
generation: through textual instructions, through content states that inherit
an input-level intervention, or through a layerwise attention channel.

For input $x$, let $h_{\ell,t}^M(x)\in\mathbb R^d$ denote the residual state
after block $\ell$ at token position $t$ under condition $M$, with $M=0$
denoting the seed instruction. We write $\tau$ for the final non-padding prompt
position, use \emph{block} and \emph{layer} interchangeably, and define the
seed-relative displacement
\begin{equation}
  \delta_{\ell,t}^M(x)=h_{\ell,t}^M(x)-h_{\ell,t}^0(x).
  \label{eq:adaptation-influence}
\end{equation}
Experiments use Gemma~2 2B-IT~\citep{riviere2024gemma2} on five-label Civil
Comments classification~\citep{borkan2019nuanced}. The primary metric is
empty-aware samples-$F_1$. A strict parser enforces the prescribed label order,
whereas a set parser evaluates only the predicted label set. Data partitions,
adapter objectives, and complete protocols are given in
Appendix~\ref{sec:diagnostic-setup}.

The insertion site specifies where an adaptation can enter the network, but it
does not identify which downstream consequence produces the gain. Likewise,
the displacement in \Cref{eq:adaptation-influence} establishes that a state
moved, but not that the movement is task-relevant. The analysis therefore
separates changes in the generated answer from changes in task information,
then tests which residual changes survive matched controls and causal transfer.

\begin{researchquestion}
Which consequences of an adaptation's insertion site remain active in the
output computation, and which measured state changes are generic effects of
changing the context?
\end{researchquestion}

\subsection{Adaptation changes both the decision and its realization}
\label{sec:dense-what}
\label{sec:dense-access}

\begin{figure}[H]
  \centering
  \begin{subfigure}[b]{0.36\linewidth}
    \centering
    \includegraphics[width=0.9\linewidth]{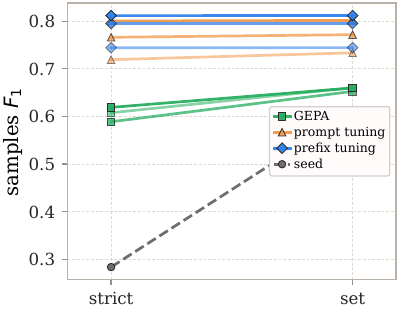}
    \caption{Label selection versus answer construction.}
    \label{fig:quality}
  \end{subfigure}\hspace{3em}
  \begin{subfigure}[b]{0.35\linewidth}
    \centering
    \includegraphics[width=0.9\linewidth]{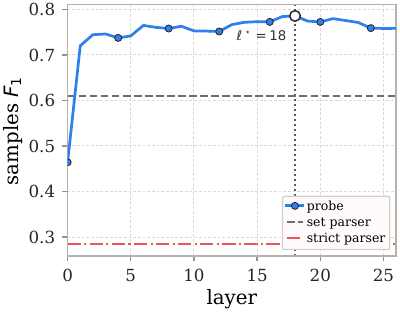}
    \caption{Seed-state label information.}
    \label{fig:probe}
  \end{subfigure}
  \caption{Behavioral gains and pre-existing task information. Ignoring label   order removes most of GEPA's strict-score improvement, while a seed-state   probe decodes the label set well before the final block.}
  \label{fig:dense-behavior}
  \vspace{-4mm}
\end{figure}

The strict evaluator combines label selection with answer construction. Under
set parsing, most of GEPA's advantage over the seed is removed, whereas prompt
and prefix tuning retain their gains (\Cref{fig:quality}). GEPA's observed gain
is therefore concentrated in the realization of the output contract in this
task, while the gains of the continuous methods remain visible in the selected
label set. The full score sweep is reported in Appendix~\ref{sec:tables}.

The seed computation already contains substantial label information. A linear
probe trained on seed states reaches most of its final accuracy by the middle
of the stack (\Cref{fig:probe}). Adaptation is therefore not applied
to a representation devoid of the task; it changes how available information
is transformed and exposed to the output computation.

\begin{figure}[H]
  \centering
  \begin{subfigure}[b]{0.59\linewidth}
    \centering
    \includegraphics[width=\linewidth,height=4.15cm,keepaspectratio]{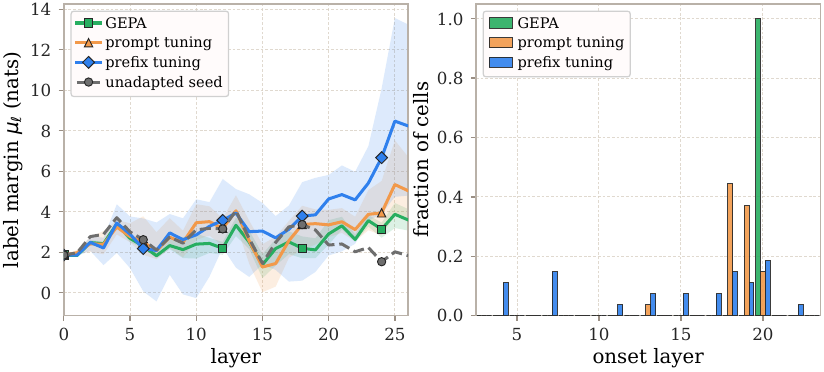}
    \caption{Depth at which adapted label margins become readable.}
    \label{fig:lens}
  \end{subfigure}\hfill
  \begin{subfigure}[b]{0.34\linewidth}
    \centering
    \includegraphics[width=\linewidth,height=4.15cm,keepaspectratio]{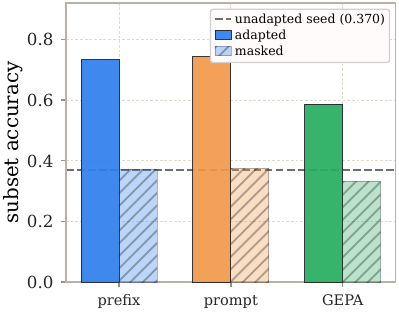}
    \caption{Adapted-context dependence.}
    \label{fig:masking}
  \end{subfigure}
  \caption{Availability and use of the adapted context. The tuned lens tracks
  the emergence of adapted label margins, while masking removes the gain when
  access to the corresponding context is blocked.}
  \label{fig:dense-readout-access}
  \vspace{-4mm}
\end{figure}

A calibrated tuned lens tracks the layerwise gold-label margin under each adapted context relative to the seed~\citep{belrose2023tunedlens}. Prefix tuning produces the earliest observed onsets in several configurations,
consistent with keys and values that are available at every block. Prompt
tuning and GEPA concentrate their onsets later in the stack
(\Cref{fig:lens}). These readouts describe availability and
masking establishes that the adapted contexts remain functionally involved. Removing attention to the optimized instruction, prompt embeddings, or prefix states returns performance to the seed level (\Cref{fig:masking}).

Together, the behavioral and readout results give an interface-level account.
GEPA's gain is concentrated in answer construction, prompt tuning introduces
an input-level influence whose label margin becomes readable later in the
stack, and prefix tuning maintains a layerwise route to the attention
computation.
Representative token-level attention profiles provide a complementary view in
Appendix~\ref{sec:attention-details}. Residual movement also follows from changing the context without optimizing it, so access alone does not make
the resulting geometry task-specific.

\subsection{Generic context produces substantial representation drift}
\label{sec:dense-which}

A length-matched padding prompt isolates the effect of adding context positions
without task optimization. It changes residual geometry, degrades probe
transfer, and lowers task performance. Padding therefore breaks any monotonic
link between distance from the seed and adaptation quality
(\Cref{fig:movement-utility} in Appendix~\ref{sec:original-shift-geometry}).

The structure of the displacement leads to the same conclusion. Early in the
stack, the dataset mean accounts for most of the shift energy under every
condition, including padding. At later blocks, the centered remainder becomes
high-rank for both adapted and padded contexts (\Cref{fig:shift-structure}). A
large shared component and a structured per-example remainder therefore also
arise when context is added without task optimization; neither signature
identifies a useful adaptation by itself.

This placebo changes the interpretation of the observational measurements.
Distance, probe transfer, mean energy, and effective rank establish that the
contexts induce different representations, but none identifies the part of the
change sufficient to reproduce adapted behavior. This is
resolved by intervening on the seed computation.

\subsection{Per-example state differences transfer adapted outputs}
\label{sec:dense-causal}

The simplest intervention treats adaptation as a shared direction. At block
$\ell$ and generation anchor $\tau$, we average the displacement over $N$
examples and add it once to the seed state:
\begin{equation}
  v_\ell^M=\frac{1}{N}\sum\nolimits_{i=1}^{N}\delta_{\ell,\tau}^M(x_i),
  \qquad
  \widetilde h_{\ell,\tau}^0(x;\lambda)
    =h_{\ell,\tau}^0(x)+\lambda v_\ell^M.
  \label{eq:mean-steering-main}
\end{equation}
Here $\lambda$ controls the magnitude of the injected direction.

\begin{wrapfigure}[11]{r}{0.42\textwidth}
  \vspace{-2.5\baselineskip}
  \centering
  \includegraphics[width=0.42\textwidth]{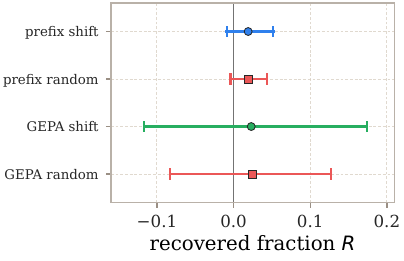}
  \caption{Dataset-mean steering and its random control.}
  \label{fig:steering}
  \vspace{-0.5\baselineskip}
\end{wrapfigure}

Across the evaluated layers and scales, the prefix mean performs like a
norm-matched random direction; the GEPA mean produces the same qualitative
comparison (\Cref{fig:steering}). The dominant average movement
in the tested intervention therefore does not reproduce the adapted behavior.
Adding the prefix mean at every decoded position recovers about $9\%$ of the
prefix gain (Appendix~\ref{sec:recurrent-mean}).

Per-example state transfer gives a different result. For source condition $S$
and target condition $T$, we add the matched-example difference
$h_{\ell,\tau}^{T}(x)-h_{\ell,\tau}^{S}(x)$ to the source state and run the
remaining source blocks unchanged~\citep{zhang2024patching}. For source,
target, and patched next-token distributions $p_S(x)$, $p_T(x)$, and
$p_{S\rightsquigarrow T}(x)$, transfer is measured by
\begin{equation}
  R_{\mathrm{KL}}(S\!\to\!T)
  =1-\frac{\sum_x\mathrm{KL}\!\left(p_T(x)\|p_{S\rightsquigarrow T}(x)\right)}
  {\sum_x\mathrm{KL}\!\left(p_T(x)\|p_S(x)\right)}.
  \label{eq:sae-transfer}
\end{equation}

\begin{wraptable}[6]{r}{0.4\textwidth}
  \vspace{-1.3em}
  \centering
  \caption{Exact transfer at block 20.}
  \vspace{-0.5em}
  \label{tab:state-transfer-main}
  % \scriptsize
  % \setlength{\tabcolsep}{3pt}
  \begin{tabular}{lc}
    \toprule
    Conditions & $R_{\mathrm{KL}}$ \\
    \midrule
    GEPA--prompt & $0.917\pm0.049$ \\
    GEPA--prefix & $0.832\pm0.056$ \\
    \bottomrule
  \end{tabular}
  \vspace{-0.5\baselineskip}
\end{wraptable}

We run this patch for GEPA paired with each continuous method, in both
directions. At a late residual state, the exact per-example difference closes
most of the source--target KL gap when aggregated across the two directions
(\Cref{tab:state-transfer-main}).
Sparse-autoencoder reconstruction and feature-restricted transfer provide a
complementary decomposition in Appendix~\ref{sec:sae-details}
\citep{lieberum2024gemmascope}.

The two interventions separate broad drift from a transferable state
difference. The tested dataset mean does not reproduce the gain, whereas one
late matched-example difference is sufficient to transfer most of the
next-token distributional difference. The exact patch uses the target state
directly; the remaining step is to produce a comparable correction from the
seed computation alone.

\subsection{Learned corrections recover adapted generation}
\label{sec:dense-predict}

For target method $M$, we learn an affine correction from the seed residual
state at the final block $L$:
\begin{equation}
  \widehat\delta_{L,t}^M=f_\phi^M(h_{L,t}^0)=W_Mh_{L,t}^0+b_M,
  \qquad
  \widetilde h_{L,t}^M=h_{L,t}^0+\widehat\delta_{L,t}^M.
  \label{eq:shift-predictor}
\end{equation}
The map has about $5.3$M parameters. Training pairs come from the adapted
model's own greedy answers: we run the seed and adapted models on each answer
and regress the seed state onto the displacement $\delta^M_{L,t}$, with state
MSE alone or together with KL divergence from the adapted output
distribution (\Cref{eq:predictor-mse,eq:predictor-mse-kl}). During free
generation, the map reads the current seed-context state and the answer
prefix produced with preceding corrections, then predicts a new correction at
every token. The adapted context and target states are absent at inference.
\citet{heyman2026steer} learn activation-dependent corrections that imitate
textual prompting with a single learned direction, and
\citet{cheng2026decomposing} study which state maps reproduce prompt-induced
task structure. Our maps are full affine corrections, fitted to all three
adaptation interfaces and evaluated by the score of complete generated
answers.

The learned corrections recover the small GEPA gap and most of the continuous
gains (\Cref{tab:predictor-main}), $81\%$ and $68\%$ of the prompt and prefix
gains with MSE and $89\%$ and $79\%$ with MSE+KL. With MSE+KL, one correction
computed at the start of the answer comes within $0.010$ of the per-token
predictor (\Cref{tab:shift-predictor}), whereas a constant bias does not beat
the seed for prompt and prefix tuning, so the gain depends on the input. 

\begin{wraptable}{r}{0.4\textwidth}
  \vspace{-0.25em}
  \centering
  % \captionsetup{justification=raggedright,singlelinecheck=false}
  \caption{Complete-answer set samples-$F_1$.}
  \vspace{-0.75em}
  \label{tab:predictor-main}
  \small
  \setlength{\tabcolsep}{1.5pt}
  \begin{tabular}{lccc}
    \toprule
    Target & Native & MSE & MSE+KL \\
    \midrule
    GEPA   & $0.678\pm.007$ & $0.683\pm.015$ & $0.681\pm.012$ \\
    Prompt & $0.777\pm.013$ & $0.746\pm.001$ & $0.759\pm.006$ \\
    Prefix & $0.805\pm.007$ & $0.744\pm.004$ & $0.765\pm.006$ \\
    \bottomrule
  \end{tabular}
  \vspace{-4mm}
\end{wraptable}
The
true shifts of GEPA and either continuous method have median cosines near
$0.2$, against $0.46$ between prompt and prefix tuning
(Appendix~\ref{sec:predictor-geometry}), matching the split in
\Cref{fig:quality}. A map from the seed block-20 state to the adapted block-25
state can even replace blocks 21 to 25, reaching $0.688$, $0.766$ and $0.770$
(Appendix~\ref{sec:early-exit}). Training details and the rank sweep are in
Appendices~\ref{sec:new-predictor-protocol} and~\ref{sec:predictor-rank}.

\paragraph{Dense-model conclusion.}
The dense experiments form a single progression. Behavioral evaluation first
separates label selection from answer construction; layerwise readouts then
show how adapted information becomes available; and the padding placebo shows
that the resulting representation drift is not specific to task improvement.
Causal transfer isolates the finer distinction: the tested dataset mean is
ineffective, while a matched-example late-state difference transfers most of
the adapted output difference. Learned affine corrections recover much of the
same behavior throughout free generation, even with one correction per
example when trained with output supervision. This establishes behavioral
transfer, but not that the maps reproduce the native adaptation mechanism.

Every dense residual state still passes through the same feed-forward
parameters. In an MoE backbone, the same state change can also alter which
experts process the token. Routing therefore exposes a discrete intermediate
decision that reveals whether adaptation changes not only the state, but also
the parameters that act on it.

% ===== End ICLR/dense_section_1 =====

% ===== Inlined from ICLR/moe_section =====
% ---------------------------------------------------------------------------
% MoE section. The appendix sections it refers to are in moe_appendix.tex.
%
% \TODO and \includegraphics are defined in dense_section_1.tex, and the
% researchquestion box in the main.tex preamble.
% ---------------------------------------------------------------------------

\section{MoE Models: When Drift Becomes a Discrete Decision}
\label{sec:moe}

\label{sec:moe-theory}

A mixture-of-experts (MoE) block sends each token to $k$ of $n_E$ experts
$f_{\ell,e}$ through a linear router $W_\ell^r$. An adapted context can therefore
change both the states processed and the parameters that process them. For
expert-block input $h_{\ell,t}$,
\begin{equation}
  g_{\ell,t} = W_\ell^{r}h_{\ell,t},\qquad
  \mathcal E_{\ell,t} = \operatorname{top-}k\bigl(g_{\ell,t}\bigr),\qquad
  y_{\ell,t} = \sum\nolimits_{e\in\mathcal E_{\ell,t}}\pi_{\ell,e}(h_{\ell,t})\,f_{\ell,e}(h_{\ell,t}),
  \label{eq:moe-layer}
\end{equation}
where $\pi_{\ell,e}$ are softmax weights over the selected experts. Through
$h_{\ell,t}$, adaptation can change the expert inputs, mixing weights and
selected set. We ask:

\begin{researchquestion}
How does each adaptation method change which experts process a token and how
evenly the load is spread, and does an expert's load tell whether the adapted
model needs it?
\end{researchquestion}

Load balancing shapes MoE training \citep{fedus2022switch,wang2024lossfree},
and frequency pruning removes rarely used experts \citep{lu2024experts}. We
test how adaptation changes this load and whether low traffic implies that an
expert is dispensable.

For expert-assignment shares $u_{\ell,e}$ and $\tilde u_{\ell,e}$ in the base
and adapted models, measured on identical comment tokens unless stated otherwise,
routing displacement is
\begin{equation}
  \mathrm{TV}_\ell=\tfrac{1}{2}\sum\nolimits_{e=1}^{n_E}\bigl|\tilde u_{\ell,e}-u_{\ell,e}\bigr|\in[0,1].
  \label{eq:moe-tv}
\end{equation}
This is the fraction of routing mass that must move to restore the base profile.
Load concentration is measured by the Gini coefficient and busiest-expert share
(Appendix~\ref{app:moe-setup}).

The task and methods follow \Cref{sec:dense}, scored by empty-aware
samples-$F_1$. Qwen3-30B-A3B-Instruct-2507 uses 48 layers, 128 experts and top-8;
gpt-oss-20b uses 24 layers, 32 experts and top-4. Both have linear gates and no
shared expert; gpt-oss adds a router bias. We use 100/200/500 virtual tokens and
500 held-out calibration rows (Appendix~\ref{app:moe-setup}).

% ===========================================================================
\subsection{Only continuous adaptation substantially reroutes the shared tokens}
\label{sec:moe-distribution}

\begin{figure}[t]
  \centering
  \includegraphics[width=\textwidth]{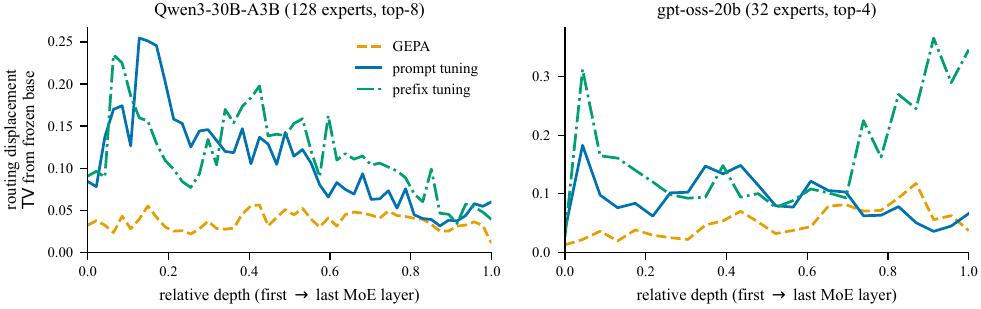}
  \caption{Routing displacement $\mathrm{TV}_\ell$ from the frozen base, layer by
  layer, for the comment tokens that all arms share; continuous arms at 500
  virtual tokens.}
  \label{fig:moe-displacement}
  \vspace{-4mm}
\end{figure}

GEPA moves only a few percent of the comment tokens' routing mass; prompt and
prefix tuning move two to four times more on average over layers
(\Cref{fig:moe-displacement}). Routing depends mostly on token identity and
less on context \citep{xue2024openmoe,olson2025semanticrouting}. GEPA changes
only the surrounding instruction, whereas learned continuous states change the
context more strongly. GEPA thus preserves most shared-token assignments, but
leaves the router's task mismatch, which router retraining removes
(Appendix~\ref{app:moe-router}). Replaying adapted routes into the base recovers
little of the task gain (Appendix~\ref{app:moe-carried}).

Continuous displacement generally falls in the late layers. The exception is
gpt-oss prefix tuning: it rises toward the experts the base uses for generated
text. GEPA's displacement also grows with depth on gpt-oss, slightly exceeding
prompt tuning in the last quarter (Appendix~\ref{app:moe-depth}).

% ===========================================================================
\subsection{The load changes with the routed sequence, not with rerouting}
\label{sec:moe-imbalance}

Adaptation moves the comment tokens between experts without concentrating their
load: every arm's Gini stays close to the base's (\Cref{tab:moe-load-full}).

\begin{table}[!b]
  \centering
  \scriptsize
  \setlength{\tabcolsep}{5pt}
  \caption{Expert load on 500 calibration rows, every budget. Gini coefficients
  and busiest-expert shares are medians over layers; dead-expert counts are means
  over layers.}
  \label{tab:moe-load-full}
  \begin{tabular}{llccccc}
    \toprule
    & & \makecell{positions\\per row} & Gini & \makecell{Gini,\\comment tokens} & \makecell{busiest\\expert (\%)} & \makecell{dead\\experts} \\
    \midrule
    \multirow{8}{*}{Qwen3-30B-A3B}
      & base         &  153 & 0.635 & 0.757 &  5.2 & 7.8 \\
      & GEPA         & 2154 & 0.729 & 0.759 &  8.2 & 7.6 \\
      & prompt, 100  &  253 & 0.583 & 0.753 &  5.0 & 3.0 \\
      & prompt, 200  &  353 & 0.610 & 0.747 &  6.1 & 4.5 \\
      & prompt, 500  &  653 & 0.620 & 0.753 &  6.6 & 3.0 \\
      & prefix, 100  &  153 & 0.640 & 0.760 &  5.8 & 8.4 \\
      & prefix, 200  &  153 & 0.647 & 0.761 &  5.6 & 8.3 \\
      & prefix, 500  &  153 & 0.648 & 0.763 &  5.5 & 9.4 \\
    \midrule
    \multirow{8}{*}{gpt-oss-20b}
      & base         &  626 & 0.513 & 0.687 & 11.2 & 0.5 \\
      & GEPA         & 1571 & 0.583 & 0.707 & 14.8 & 0.5 \\
      & prompt, 100  &  313 & 0.563 & 0.693 & 15.8 & 1.0 \\
      & prompt, 200  &  413 & 0.607 & 0.689 & 17.9 & 0.8 \\
      & prompt, 500  &  713 & 0.661 & 0.711 & 19.3 & 1.2 \\
      & prefix, 100  &  213 & 0.584 & 0.682 & 15.6 & 2.2 \\
      & prefix, 200  &  213 & 0.572 & 0.685 & 14.8 & 2.0 \\
      & prefix, 500  &  213 & 0.591 & 0.661 & 17.0 & 1.6 \\
    \bottomrule
  \end{tabular}
  \vspace{-4mm}
\end{table}

% \newpage
\begin{wraptable}{r}{0.30\textwidth}
  \centering
  \small
  \vspace{-4mm}
  \setlength{\tabcolsep}{4pt}
  \caption{gpt-oss test $F_1$, 2000 rows. Reasoning LR: answer-only for prompts,
  $3\cdot10^{-4}$ for prefixes.}
  \vspace{-2mm}
  \label{tab:moe-gptoss-reasoning}
  \begin{tabular}{lcc}
    \toprule
    arm & \makecell{answer\\only} & \makecell{with\\reasoning} \\
    \midrule
    prompt, 100 & 0.784 & 0.482 \\
    prompt, 500 & 0.799 & 0.775 \\
    prefix, 100 & 0.819 & 0.789 \\
    prefix, 500 & 0.820 & 0.764 \\
    \bottomrule
  \end{tabular}
\end{wraptable}

The full-sequence load changes mainly through the tokens adaptation adds or
removes. GEPA's long instruction multiplies routed positions; Qwen's virtual
tokens spread traffic to nearly idle experts. On gpt-oss the tuned arms instead
concentrate the load by omitting the base's reasoning.

\paragraph{The adapted gpt-oss answers directly.}
\label{sec:moe-gptoss-mode}
The adapters learn to answer directly. Including the base's analysis in
training, masked from the loss, preserves reasoning but lowers task $F_1$
across the tested arms (\Cref{tab:moe-gptoss-reasoning}). Forcing direct answers
also lowers the frozen base's $F_1$: $0.629$ to $0.495$ with the seed instruction
and $0.755$ to $0.723$ with GEPA. Answer-only training benefits the adapters;
forcing that mode on the frozen base does not.

% ===========================================================================
\subsection{Frequency pruning removes experts the adapted model needs}
\label{sec:moe-pruning}
% The least-used half of each layer carries 2--13 percent of routing mass, and
% three quarters 18--40 percent, averaged over layers. We mask the router logits
% of these experts, ranked on each arm's own calibration map.

A concentrated load leaves many experts nearly idle: on this task the least-used half of a layer's experts carries, on average over layers, only 2 to 13 percent of the routing mass, and the least-used three quarters 18 to 40 percent.
Such experts are the natural candidates for removal.
We prune them by masking, in every layer, the router logits of the least-used half or three quarters of the experts, ranked on the arm's own calibration map.
Removing little routing mass can still destroy readable answers
(\Cref{tab:moe-pruning}; full sweep in Appendix~\ref{app:moe-pruning}).

\begin{table}[t]
  \centering
  \scriptsize
  \setlength{\tabcolsep}{5pt}
 \caption{Set-level $F_1$ on 2000 test rows after pruning each layer's least-used 50\% or 75\% of experts, with unreadable-answer rates; $F_1$ after pruning only super experts, alongside those used by each arm. Token budgets (Qwen/gpt-oss): 512/1024 for unpruned and frequency-pruned models, 64/1024 for super-expert pruning; unreadable answers in the latter count as empty label sets. Empty-label floors: 0.330 for frequency pruning and 0.328 for super-expert pruning, using different test samples.}
  \label{tab:moe-pruning}
  \setlength{\tabcolsep}{4pt}
  \begin{tabular}{llccccclc}
    \toprule
    & & \multicolumn{3}{c}{$F_1$} & \multicolumn{2}{c}{unreadable (\%)} & \multicolumn{2}{c}{super experts} \\
    \cmidrule(lr){3-5}\cmidrule(lr){6-7}\cmidrule(lr){8-9}
    & & original & $-50\%$ & $-75\%$ & $-50\%$ & $-75\%$ & kept & pruned \\
    \midrule
    \multirow{8}{*}{Qwen3-30B-A3B}
      & base         & 0.466 & 0.575 & 0.330 & 0.0 & 100.0 & all three & 0.328 \\
      & GEPA         & 0.728 & 0.728 & 0.330 & 0.0 & 100.0 & all three & 0.328 \\
      & prompt, 100  & 0.833 & 0.814 & 0.440 & 0.0 & 0.2 & L2\,E92, L3\,E82 & 0.409 \\
      & prompt, 200  & 0.816 & 0.330 & 0.330 & 100.0 & 100.0 & all three & 0.328 \\
      & prompt, 500  & 0.828 & 0.812 & 0.450 & 0.0 & 0.0 & all three & 0.328 \\
      & prefix, 100  & 0.836 & 0.815 & 0.383 & 0.0 & 0.3 & all three & 0.328 \\
      & prefix, 200  & 0.802 & 0.611 & 0.330 & 0.0 & 100.0 & L2\,E92, L3\,E82 & 0.328 \\
      & prefix, 500  & 0.805 & 0.330 & 0.330 & 100.0 & 100.0 & L3\,E82 & 0.328 \\
    \midrule
    \multirow{8}{*}{gpt-oss-20b}
      & base         & 0.631 & 0.565 & 0.388 & 19.6 & 34.5 & both & 0.660 \\
      & GEPA         & 0.755 & 0.746 & 0.422 & 0.3 & 64.4 & both & 0.754 \\
      & prompt, 100  & 0.788 & 0.502 & 0.330 & 42.4 & 98.2 & both & 0.726 \\
      & prompt, 200  & 0.788 & 0.584 & 0.343 & 26.2 & 84.3 & both & 0.784 \\
      & prompt, 500  & 0.800 & 0.607 & 0.334 & 14.3 & 96.8 & both & 0.768 \\
      & prefix, 100  & 0.821 & 0.736 & 0.528 & 0.0 & 0.9 & none & 0.818 \\
      & prefix, 200  & 0.826 & 0.793 & 0.486 & 0.0 & 3.0 & both & 0.814 \\
      & prefix, 500  & 0.818 & 0.797 & 0.405 & 0.0 & 1.1 & none & 0.823 \\
    \bottomrule
  \end{tabular}
  \vspace{-4mm}
\end{table}

% ===========================================================================
\subsection{Rarely used super experts decide whether an adapted model survives}
\label{sec:moe-super}

Load is not the only way in which experts differ. Language models send much of
their attention to a few initial tokens with little content, the attention sinks
\citep{xiao2024streaming}, which are created by massive activations, a few
hidden-state entries orders of magnitude larger than the rest
\citep{sun2024massive,gu2025sink}. In MoE models these are written by a handful
of super experts with rare, extreme outputs, whose removal collapses the Qwen
model studied by \citet{su2026superexperts}. We recover experts 68, 92 and 82
of layers 1 to 3 on Qwen and identify expert 5 of layers 6 and 17 on gpt-oss
(Appendix~\ref{app:moe-super}). Unlike a
Qwen head, every head of gpt-oss has a learned sink logit that lets it attend to
no token \citep{openai2025gptoss}.

Qwen's router sends its super experts at most about the uniform share of tokens
(0.1 to 1.3 times) and gpt-oss's about three times that share (2.9 and 3.0
times); in every adapted gpt-oss map expert 5 of layer 17 keeps 3.5 to 7.3
times it, so frequency masks remove all of Qwen's but never both of gpt-oss's.
Every Qwen cell whose mask removes all three loses
every readable answer. In the two $-50\%$ rescue controls, masks of the same
size that spare them restore readable answers, with $F_1$ of $0.803$ for
prompt-200 and $0.768$ for prefix-500, against $0.816$ and $0.805$ intact
(Appendix~\ref{app:moe-pruning}).

\begin{figure}[t]
  \centering
  \includegraphics[width=\textwidth]{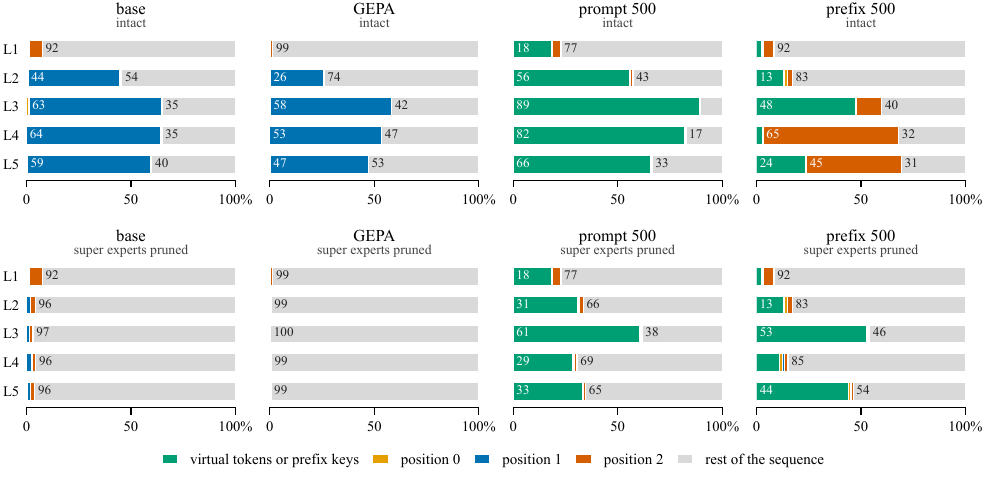}
  \vspace{-6mm}
  \caption{Share of attention by layer on Qwen, intact and with the super experts
  pruned.}
  \label{fig:moe-attention}
  \vspace{-4mm}
\end{figure}

\begin{wraptable}{r}{0.5\textwidth}
  \centering
  \small
  \vspace{-10mm}
  \setlength{\tabcolsep}{4pt}
  \caption{Largest output of each super expert over 200 rows.}
  \label{tab:moe-super-mag}
  \vspace{-4mm}
  \begin{tabular}{lccccc}
    \toprule
    & \multicolumn{3}{c}{Qwen3-30B-A3B} & \multicolumn{2}{c}{gpt-oss-20b} \\
    \cmidrule(lr){2-4}\cmidrule(lr){5-6}
    & L1\,E68 & L2\,E92 & L3\,E82 & L6\,E5 & L17\,E5 \\
    \midrule
    base        & 422  & 800  & 458 & 43\,000 & 52\,700 \\
    GEPA        & 422  & 800  & 458 & 43\,000 & 52\,700 \\
    prompt, 100 & 11.1 & 250  & 572 & 43\,000 & 50\,700 \\
    prompt, 200 & 470  & 764  & 472 & 43\,000 & 51\,500 \\
    prompt, 500 & 1000 & 952  & 572 & 43\,000 & 46\,900 \\
    prefix, 100 & 592  & 776  & 464 & 290     & 1\,520  \\
    prefix, 200 & 0.65 & 188  & 486 & 43\,000 & 47\,900 \\
    prefix, 500 & 0.55 & 16.1 & 520 & 189     & 1\,990  \\
    \bottomrule
  \end{tabular}
  \vspace{-6mm}
\end{wraptable}

Adaptation changes dependence on these experts (\Cref{tab:moe-super-mag}).
GEPA leaves their magnitudes unchanged; prompt tuning moves their peak onto a
virtual token (\Cref{fig:moe-attention}); prefix tuning can suppress them by
three orders of magnitude, depending on budget, learning rate and seed
(Appendix~\ref{app:moe-super}).

Removing the super experts destroys readable answers on all selected Qwen
arms except prompt-100 (\Cref{tab:moe-pruning}).
On gpt-oss, score changes are smaller and depend on the arm. Prompt-100 falls
from $0.784$ to $0.726$ in the super-expert harness. The prefix arms that suppress both experts stay close
to their intact scores. Attention shifts to the learned sink and the sequence
(\Cref{fig:moe-attention-gptoss}), consistent with reduced dependence on
token-based sinks.

% \WFclear
\paragraph{MoE conclusion.}
Adaptation reroutes an MoE model where it acts: GEPA keeps most expert
assignments of the shared tokens, prompt and prefix tuning reassign more of them.
Load statistics and pruning masks built from them are specific to each
adapter, and load does not show which experts the model needs: a few rarely used
super experts decide whether an arm survives pruning.

% ===== End ICLR/moe_section =====

% \newpage
\section{Conclusion}
\label{sec:conclusion}

Prompt-based adaptation methods differ less in how much they change a frozen
model than in where they act, with practical consequences for evaluation,
interpretation and deployment. Under set parsing most of GEPA's gain
disappears while the continuous methods keep theirs, so an optimized prompt
should be scored both ways before its gain is credited to task knowledge. A
length-matched padding prompt reproduces the distance, probe, mean-energy and
rank signatures of adaptation without improving the task, so representation
metrics need such a placebo as a reference. An affine map from the seed state
recovers most of the continuous gains without the adapted context and can even
replace the last blocks, which makes such maps a practical probe of what an
adaptation computes, although they need not reproduce the native mechanism.

In MoE models, retraining the router alone recovers most of the adaptation
gain, and the choice of adaptation is also a serving decision. Text-level
optimization keeps most of the routing of shared tokens but lengthens every request,
whereas continuous prompts reroute many of them, so expert placement, caches
and pruning masks tuned on the base should be recomputed per adapter. Pruning
by traffic alone is unsafe, since a few rarely used super experts decide
whether a pruned model still answers and the adaptation decides whether it
needs them.

\clearpage
\bibliography{refs}
\bibliographystyle{plainnat}

\clearpage
\appendix

\section{Numerical Results}
\label{sec:tables}
Let $\Delta_{\mathrm{str}}$ and $\Delta_{\mathrm{set}}$ denote the gain over the
seed under order-sensitive and order-invariant parsing. We report the
fraction of the former gain that disappears under the latter evaluator:
\begin{equation}
  \rho =
  \frac{\Delta_{\mathrm{str}}-\Delta_{\mathrm{set}}}
       {\Delta_{\mathrm{str}}}.
  \label{eq:parsegain}
\end{equation}
Both the seed and adapted scores change with the parser. Thus $\rho$ is an
operational comparison of evaluation rules, not a causal decomposition of
learning into formatting and label prediction. For example, at $n=1000$,
prompt tuning scores $0.8003$ and $0.8017$, but its gains over the seed are
$0.5166$ and $0.1925$ under the two rules.

For a residual intervention $\lambda v_\ell^M$, the recovered fraction of the
target task-score gap is
\begin{equation}
  R_{\mathrm{task}}^M(\ell,\lambda)=
  \frac{F_{\ell,\lambda v_\ell^M}(\mathcal C_0)-F(\mathcal C_0)}
       {F(\mathcal C_M)-F(\mathcal C_0)},
  \label{eq:recovery}
\end{equation}
where $F$ is the task metric and $\mathcal C_0$ and $\mathcal C_M$ are the
seed and adapted conditions.

\begin{table}[ht]
  \centering
  \caption{Held-out test performance, $3000$ rows, empty-aware samples $F_1$
  under both parsers, used in \Cref{fig:quality}. The
  unadapted seed scores $0.2837$ strict and $0.6092$ set. The parser-dependent
  gain reduction
  $\rho$ of \Cref{eq:parsegain} is $0.85$ for the discrete branch. GEPA entries
  average over three optimizer seeds. Continuous entries average over
  $m\in\{100,200,500\}$ and three seeds.}
  \label{tab:quality}
  \begin{tabular}{lcccccc}
    \toprule
    & \multicolumn{2}{c}{discrete (GEPA)} & \multicolumn{2}{c}{prompt tuning} & \multicolumn{2}{c}{prefix tuning} \\
    \cmidrule(lr){2-3}\cmidrule(lr){4-5}\cmidrule(lr){6-7}
    $n$ & strict & set & strict & set & strict & set \\
    \midrule
    $200$  & 0.6077 & 0.6604 & 0.7192 & 0.7336 & 0.7443 & 0.7445 \\
    $500$  & 0.5888 & 0.6526 & 0.7662 & 0.7717 & 0.7951 & 0.7952 \\
    $1000$ & 0.6190 & 0.6598 & 0.8003 & 0.8017 & 0.8117 & 0.8118 \\
    \bottomrule
  \end{tabular}
\end{table}

\begin{table}[ht]
  \centering
  \caption{Recovered fraction $R_{\mathrm{task}}$ of \Cref{eq:recovery}, aggregated over $12$
  cells per branch, against a norm-matched random direction injected at the same
  site, used in \Cref{fig:steering}.}
  \label{tab:steering}
  \begin{tabular}{lcc}
    \toprule
    branch & $R$ [95\% CI] & norm-matched random \\
    \midrule
    prefix tuning ($m=500$) & $+0.019$ $[-0.009, +0.052]$ & $+0.019$ $[-0.004, +0.043]$ \\
    discrete (GEPA)         & $+0.023$ $[-0.117, +0.174]$ & $+0.024$ $[-0.083, +0.127]$ \\
    \bottomrule
  \end{tabular}
\end{table}

\section{Additional Dense Diagnostics}
\label{sec:dense-appendix}
This section reports the diagnostics of \Cref{sec:dense-what,sec:dense-which}
and the adapters, partitions and interventions they use.

\subsection{Adaptation Objectives and Insertion Sites}
\label{sec:adaptation-details}

Each method optimizes an auxiliary object $z$ with the backbone $\theta$
frozen. For dataset $\mathcal D$ and method-specific feasible set $\mathcal Z_M$,
\begin{equation}
  z_M^\star \in \arg\min_{z\in\mathcal Z_M}\mathcal J_M(z;\mathcal D),
  \label{eq:adaptation-objective-main}
\end{equation}
where $\mathcal J_M$ is the autoregressive sequence loss for continuous
methods and the negative black-box evaluation score for GEPA. GEPA searches
textual instructions through reflection and evolutionary
selection~\citep{agrawal2026gepa}. Prompt tuning learns input embeddings
$P\in\mathbb R^{m\times d}$~\citep{lester2021prompt}, and prefix tuning learns
keys and values $\{K_\ell^P,V_\ell^P\}_{\ell=1}^L$ for every attention
block~\citep{li2021prefix}. In the formal recurrence, blocks are indexed
$1,\ldots,L$, with $H_0$ the input state and $H_L$ the final block output.
Dense experiments use zero-based block labels $0,\ldots,L-1$:
$h_{\ell,\tau}$ is token $\tau$ in $H_{\ell+1}$, so Gemma block 25 outputs $H_{26}$.
The insertion sites are
\begin{equation}
  \begin{aligned}
  H_0^M(x) &=
  \begin{cases}
    E\!\left([z_{\mathrm{GEPA}};\mathcal C(x)]\right), & M=\mathrm{GEPA},\\
    [P;E(\mathcal C(x))], & M=\mathrm{prompt},\\
    E(\mathcal C(x)), & M=\mathrm{prefix},
  \end{cases}\\[4pt]
  \operatorname{Attn}_\ell^{\mathrm{prefix}}
    &= \operatorname{softmax}\!\left(
       \frac{Q_\ell[K_\ell^P;K_\ell]^{\top}}{\sqrt{d_k}}+\mathcal M_\ell
       \right)[V_\ell^P;V_\ell].
  \end{aligned}
  \label{eq:adaptation-sites}
\end{equation}
Here $\mathcal C(x)$ is the common scaffold, $Q_\ell$, $K_\ell$ and $V_\ell$
are the content projections, and $\mathcal M_\ell$ extends the causal mask
with unmasked prefix columns. Text and prompt embeddings enter at the
input, while prefixes supply learned keys and values at every layer. These
parameterizations provide different ways to condition the frozen model,
as studied in theoretical analyses of continuous
prompting~\citep{oymak2023attention,petrov2023prompting}.

For the gradient-trained arms and target sequence $y=(y_1,\ldots,y_T)$, the loss in
\Cref{eq:adaptation-objective-main} is the autoregressive negative
log-likelihood
\begin{equation}
  \mathcal L_{\mathrm{NLL}}(x,y;z)
    =-\sum_{t=1}^{T}\log p_\theta
      \!\left(y_t\mid y_{<t},x;z\right).
  \label{eq:adapter-nll}
\end{equation}
Prompt tuning optimizes $P\in\mathbb R^{m\times d}$ and concatenates it with the
embedded textual scaffold only once. Prefix tuning optimizes a separate block
of $m$ keys and values for every attention layer. These are concatenated
with the content keys and values as in \Cref{eq:adaptation-sites}, but do not
enter the residual stream as token positions. GEPA instead treats the task score
as a black-box objective and may rewrite only the natural-language instruction.
The label list, input wrapper, and output contract remain fixed in every arm.

\subsection{Diagnostic Setup}
\label{sec:diagnostic-setup}

The backbone is \texttt{google/gemma-2-2b-it}, with $26$ transformer blocks and
hidden size $2304$. Civil Comments scores are thresholded at $0.5$ for five
non-exclusive labels: \texttt{toxicity}, \texttt{obscene}, \texttt{threat},
\texttt{insult}, and \texttt{identity\_attack}. Empty-aware samples-$F_1$
assigns one when the prediction and target sets are both empty. About one
third of each split has an empty target set.

The adaptation, probe, intervention and $3000$-example test
partitions are disjoint. The performance sweep uses optimizer pools
$n\in\{200,500,1000\}$ and three seeds. GEPA has a budget of at most
$10^4$ metric calls and uses \texttt{gpt-5.6-luna-pro} as its reflector.
Prompt and prefix tuning use the same optimizer pools and
$m\in\{20,100,200,500\}$ virtual tokens. \Cref{tab:quality} averages over
$m\in\{100,200,500\}$. The final test set is not used to train probes or
select adapted prompts.

The length-matched padded prompt $\mathcal C_{\mathrm{pad}}$ adds context
without adaptation. We compare an internal measurement under the adapted
and padded contexts before attributing it to adaptation. A change shared
with padding can arise from adding context positions alone.

In \Cref{tab:quality}, the continuous methods improve monotonically with
optimizer pool size, while GEPA does not. Averaged over the three pool
sizes, the prefix lead over GEPA is $0.179$ under order-sensitive parsing
and $0.126$ under set parsing. Ignoring output order reduces this measured
gap by about $30\%$, an evaluator-dependent comparison rather than a causal
decomposition of the gain. Appendix~\ref{sec:gepa-prompts} gives an example
of the optimized GEPA instruction from this sweep.

\paragraph{Mean displacement and the padding control.}
For each method, the one-shot intervention uses the mean displacement
$v_\ell^M$ at the generation anchor $\tau$ defined in
\Cref{eq:mean-steering-main}. Subsequent positions receive no direct
intervention in this setting.
We measure the fraction of displacement energy captured by the mean,
\begin{equation}
  \eta_\ell^M=\frac{\|v_\ell^M\|_2^2}
  {N^{-1}\sum_i\|\delta_{\ell,\tau}^M(x_i)\|_2^2}.
  \label{eq:eta}
\end{equation}
At layer 6, this fraction is $0.961$ for GEPA, $0.981$ for prefix tuning,
$0.967$ for prompt tuning, and $0.948$ for padding. At layer 18 the respective
values are $0.636$, $0.676$, $0.569$, and $0.484$. A large shared displacement
therefore also occurs under a context change that does not improve the task.
The one-shot steering sweep covers four layers and three seeds per branch.
The reported relative-norm scale settings are $\{0.05,0.1,0.2\}$, measured
against the state norm at the injection site. These settings specify the
perturbation size, while $\lambda$ in \Cref{eq:mean-steering-main} denotes
the multiplier of the mean displacement. Norm-matched random directions
are injected at the same sites. \Cref{tab:steering} reports the aggregated recovery. The prefix
interval is narrow around a small effect, while the GEPA interval is wide
because its target gap is small. The reported recoveries for both branches
are similar to those of their random controls.

\subsection{Tuned Lens and Linear Probes}
\label{app:lens}

We measure label decodability with a linear probe~\citep{hewitt2019probes}
and the output distribution accessible through a calibrated tuned
lens~\citep{belrose2023tunedlens}. Their readouts are
\begin{equation}
  p_\ell^{\mathrm{TL}}(x)=\operatorname{softmax}\!\left(W_U\operatorname{LN}(A_\ell h_{\ell,\tau}(x)+b_\ell)\right),
  \qquad
  \widehat y_{c,\ell}(x)=\sigma\!\left(w_{c,\ell}^{\top}h_{\ell,\tau}(x)+\beta_{c,\ell}\right).
  \label{eq:dense-readouts}
\end{equation}
At each layer, the affine lens translator minimizes the expected KL from
the model's final output distribution to $p_\ell^{\mathrm{TL}}$.
The regularized logistic probes use $8000$ training and $2000$ validation
examples, extracting the residual state at the last prompt position.
These auxiliary readouts measure available information, without identifying
the computation used by the model's own output head.

For the tuned lens, we measure the margin between present and absent labels,
\begin{equation}
  \mu_\ell(x)=\frac{1}{|G|}\sum_{c\in G}\log p_\ell^{\mathrm{TL}}(c)
  -\frac{1}{|\bar G|}\sum_{c\in\bar G}\log p_\ell^{\mathrm{TL}}(c),
  \label{eq:margin}
\end{equation}
where $G$ is the gold label set. We average only over examples with
$0<|G|<5$, for which both terms are defined. GEPA remains above the seed
from layer 20 in every evaluated configuration. Prompt tuning crosses at
about layer 18, while prefix tuning crosses between layers 4 and 20
(\Cref{fig:lens}).

Under the seed prompt, probe validation samples-$F_1$ rises from $0.464$ at
the first layer to about $0.75$ by layer 10 and peaks at $0.7855$ in layer 18
(\Cref{fig:probe}). A seed-trained probe scores between $0.77$ and $0.83$
on seed states but between $0.16$ and $0.73$ on adapted states. Retraining
within each adapted stream restores scores between $0.74$ and $0.80$.
Length-matched padding similarly degrades probe transfer. Failed transfer
therefore does not, by itself, identify an adaptation-specific mechanism.

\subsection{Attention to the Adapted Context}
\label{sec:attention-details}

For head-averaged attention $\alpha_\ell(t,j)$ from the read position $t$,
we sum over virtual-token positions to measure attention to continuous
contexts. For a textual instruction, we plot cumulative mass
$M_\ell(k)=\sum_{j\leq k}\alpha_\ell(t,j)$ along its positions.
The prompt diagnostic uses $m=20$, $n=20000$, and seed 43. Its virtual-token
mass peaks at $0.252$ in layer 7 and falls to $0.021$ at the final layer.
The prefix diagnostic uses $m=500$, $n=20000$, and seed 44. Its virtual-token
mass rises from $0.523$ to $0.940$ and remains between $0.90$ and $0.95$
during generation. Its attention mass on the text instruction falls to
$0.033$. The models differ in seed and capacity, so these values
do not isolate an effect of adaptation method.

\begin{figure}[t]
  \centering
  \includegraphics[width=0.82\linewidth]{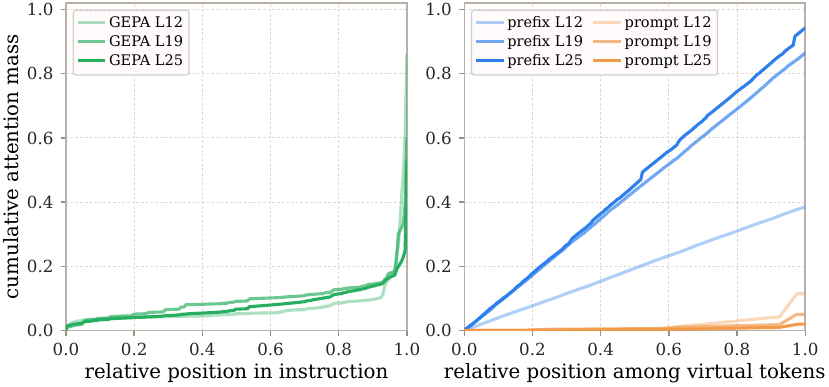}
  \caption{Representative access profiles for the adapted contexts. Curves
  show cumulative attention within the GEPA instruction (left) and learned
  continuous contexts (right) at selected blocks.}
  \label{fig:cummass}
\end{figure}

The optimized GEPA instruction contains about $1850$ tokens. At layers 12,
19 and 25, cumulative attention grows mainly over the opening task statement
and closing format rule, with little growth over the intervening label
criteria (\Cref{fig:cummass}). These profiles are descriptive. Masking
access to either continuous context returns subset accuracy to the zero-shot
level, providing the intervention test reported in the main text.
For GEPA, masking the instruction lowers subset accuracy slightly below
the seed level (\Cref{fig:masking}).

\subsection{Mean Shifts and Cosine Geometry}
\label{sec:original-shift-geometry}

\begin{figure}[t]
  \centering
  \begin{subfigure}[b]{0.38\linewidth}
    \centering
    \includegraphics[width=\linewidth,height=3.65cm,keepaspectratio]{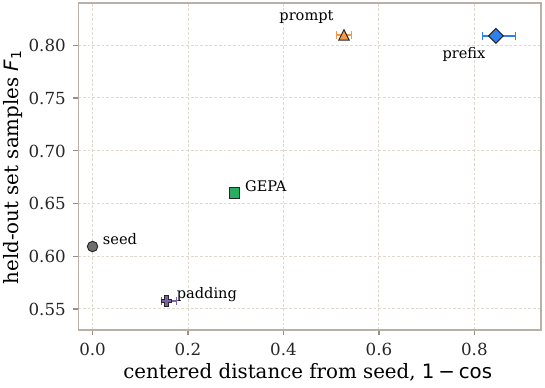}
    \caption{Movement does not order task quality.}
    \label{fig:movement-utility}
  \end{subfigure}\hfill
  \begin{subfigure}[b]{0.6\linewidth}
    \centering
    \includegraphics[width=\linewidth,height=3.65cm,keepaspectratio]{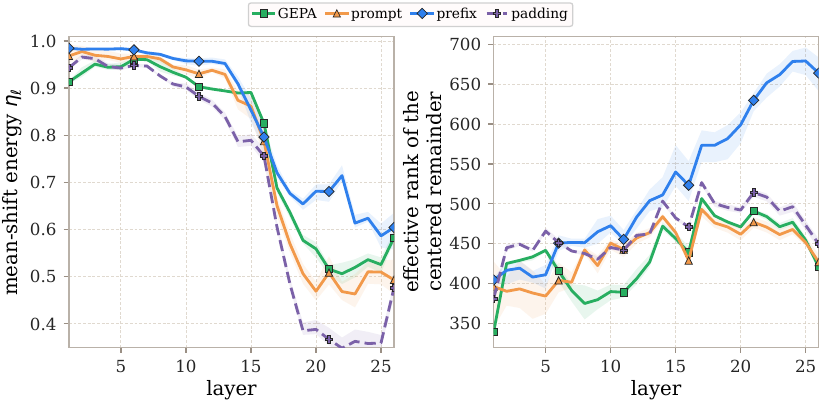}
    \caption{Mean energy and rank under adaptation and padding.}
    \label{fig:shift-structure}
  \end{subfigure}
  \caption{A matched context placebo separates representation drift from task
  utility. Padding reproduces prominent geometric signatures without improving
  the task.}
  \label{fig:dense-drift-controls}
\end{figure}

We analyze the diagnostic states at the final prompt position on
$N=1000$ fixed examples. These measurements concern the observed
seed-relative displacements, rather than the learned predictor matrices
in Appendix~\ref{sec:predictor-geometry}. After subtracting the dataset mean
in \Cref{eq:eta}, we measure the effective rank of the remaining variation.
With $d=2304$ and $N=1000$, the centered matrix has rank at most $999$.

At layer 18, the centered remainder has effective rank between $476$ and
$573$ across adaptation methods, compared with $500$ for padding
(\Cref{fig:shift-structure}). Both the large mean component and the broad
remaining variation therefore occur under a context change that does not
improve the task.

The pairwise mean can hide how shift directions vary across inputs. We retain
this variation with
\begin{equation}
  \kappa_\ell^{A,B}(x)
  = \frac{\langle\delta_{\ell,\tau}^A(x),\delta_{\ell,\tau}^B(x)\rangle}
  {\|\delta_{\ell,\tau}^A(x)\|_2\,\|\delta_{\ell,\tau}^B(x)\|_2}.
  \label{eq:pairwise-shift-cosine}
\end{equation}

\begin{figure}[t]
  \centering
  \includegraphics[width=\linewidth]{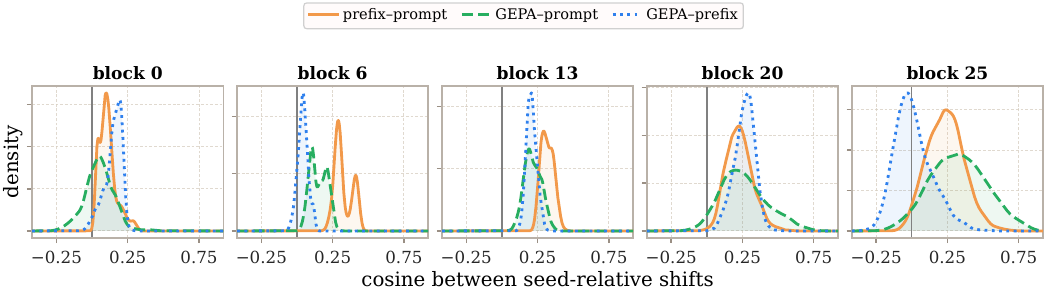}
  \caption{Per-example cosine between seed-relative shifts from two adaptation
  methods. Each panel pools 1000 fixed examples. Early-layer shifts have small norms, so their directions are less
  stable.}
  \label{fig:shift-cosine-distributions}
\end{figure}

At block 25, mean per-example cosine is $0.011$ for GEPA and prefix,
$0.238$ for prefix and prompt, and $0.327$ for GEPA and prompt. The last
pair has the broadest distribution (\Cref{fig:shift-cosine-distributions}).
These means use the diagnostic adapters of
Appendix~\ref{sec:diagnostic-setup}. Appendix~\ref{sec:predictor-geometry}
reports the corresponding medians for the adapters of the predictor
experiment.

\begin{table}[ht]
  \centering
  \caption{Mean per-example cosine similarity with the seed-prompt residual at
  layer 18 over three seeds. Centered states have their dataset mean removed.}
  \label{tab:cosine}
  \begin{tabular}{lcc}
    \toprule
    condition & raw & centered \\
    \midrule
    GEPA                         & 0.902 & 0.702 \\
    prompt tuning ($m=500$)      & 0.830 & 0.473 \\
    prefix tuning ($m=500$)      & 0.497 & 0.155 \\
    $\mathcal{C}_{\mathrm{pad}}$ & 0.961 & 0.845 \\
    \bottomrule
  \end{tabular}
\end{table}

Subtracting each condition's dataset mean reduces its cosine similarity
with the seed (\Cref{tab:cosine}). GEPA remains more aligned with the seed
than either continuous method. Prefix tuning has the lowest centered cosine,
$0.155$, while padding remains much closer at $0.845$.

\Cref{fig:movement-utility} compares this distance with task quality.
Padding stays close to the seed but performs worse. GEPA improves while
moving less than either continuous method. The geometric distance alone
therefore does not identify a useful intervention.

\subsection{SAE Fidelity and Cross-Method State Patching}
\label{sec:sae-details}

We apply Gemma Scope SAEs~\citep{lieberum2024gemmascope} to residual states
at real token positions.
Learned input embeddings and prefix keys and values do not share these
residual coordinates. On $500$ test examples, replacing one selected state
by its SAE reconstruction changes samples-$F_1$ by at most $0.014$.
Output agreement, averaged over ten evaluated layers, ranges from $0.873$
to $0.975$ across the evaluated conditions. This checks the reconstruction's
behavioral fidelity before using it to represent a displacement.

We also measure how well the SAE reconstructs the displacement itself.
For method $M$ relative to the seed, the dense displacement, decoded
displacement and recovered energy are
\begin{equation}
  \begin{aligned}
    \delta_{\ell,\tau}^M(x)
      &= h_{\ell,\tau}^M(x)-h_{\ell,\tau}^0(x), \\
    \widehat{\delta}_{\ell,\tau}^M(x)
      &= \operatorname{Dec}_\ell\!\left(\operatorname{Enc}_\ell(h_{\ell,\tau}^M(x))\right)
       - \operatorname{Dec}_\ell\!\left(\operatorname{Enc}_\ell(h_{\ell,\tau}^0(x))\right), \\
    q_\ell^M
      &= 1-
      \frac{\sum_x\|\delta_{\ell,\tau}^M(x)-\widehat{\delta}_{\ell,\tau}^M(x)\|_2^2}
      {\sum_x\|\delta_{\ell,\tau}^M(x)\|_2^2}.
  \end{aligned}
  \label{eq:sae-shift}
\end{equation}
In the matched $n=1000$, $m=20$ comparison, $q_{25}$ is $0.421$ for prompt
tuning, $0.582$ for prefix tuning and $0.616$ for GEPA. Recovered energy can
be negative in early layers, where reconstruction error exceeds the energy
of the displacement. We therefore restrict this analysis in SAE coordinates
to late layers. Recovered energy measures state reconstruction, while the
following intervention measures transfer of the output distribution.

The following state-transfer experiment uses a separate adapter set:
continuous adapters trained with $n=20000,m=20$ and GEPA prompts trained
with $n=1000$, rather than the matched set used for recovered energy above.
For a source condition $S$ and a target condition $T$, the exact patch adds
$h_{\ell,\tau}^{T}(x)-h_{\ell,\tau}^{S}(x)$ to the source state, as defined in
the main text. Its SAE counterpart is
\begin{equation}
  \widehat\delta_{\ell,\tau}^{S\to T}(x)
  =\operatorname{Dec}_\ell\!\left(\operatorname{Enc}_\ell(h_{\ell,\tau}^{T}(x))\right)
  -\operatorname{Dec}_\ell\!\left(\operatorname{Enc}_\ell(h_{\ell,\tau}^{S}(x))\right).
  \label{eq:sae-input-dependent-patch}
\end{equation}
The patch is applied at the final non-padding prompt position, and the
remaining source blocks run unchanged~\citep{zhang2024patching}. It therefore
concerns the first output distribution, before either model has generated
an answer. We report the fraction of the next-token KL gap closed by each
patch using $R_{\mathrm{KL}}$ from \Cref{eq:sae-transfer}.
The bidirectional GEPA/prompt comparison recovers $0.917\pm0.049$ with exact
states and $0.868\pm0.012$ through the SAE. The corresponding GEPA/prefix
values are $0.832\pm0.056$ and $0.708\pm0.084$. They describe state transfer
between the selected models, rather than a controlled estimate of
adaptation-method effects.

\begin{figure}[t]
  \centering
  \includegraphics[width=0.92\linewidth]{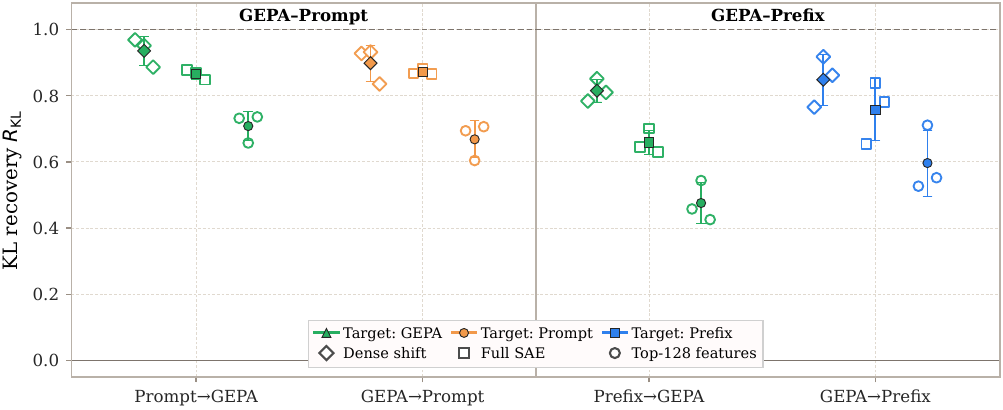}
  \caption{Exact-state, SAE-reconstructed, and Top-128 feature transfer. The
  Top-128 subset is selected by displacement magnitude on disjoint calibration
  examples.}
  \label{fig:sae-causal-transfer-full}
\end{figure}

Selecting the $128$ most changed SAE features on $32$ disjoint calibration
examples recovers $0.688$ of the GEPA/prompt gap and $0.536$ of the GEPA/prefix
gap. This feature subset is selected by change magnitude. Its transfer score
does not identify the semantics of the selected features
(\Cref{fig:sae-causal-transfer-full}).

\section{Shift Predictor Details}
\label{sec:predictor-details}

\subsection{MSE and MSE+KL on 3000 Test Examples}
\label{sec:new-predictor-protocol}

The new evaluation compares state MSE with MSE plus an output KL term at the
final block, in two regimes. Repredict computes a new shift at each generated
position. Fixed computes one shift at the generation anchor and reuses it.
Fixed-regime details and a low-rank ablation are in
Appendix~\ref{sec:predictor-rank}.
Appendix~\ref{sec:legacy-predictor} reports an earlier experiment with
additional SAE losses.
The manual seed-model samples-$F_1$ on these $3000$ examples is $0.617$.
Block 25 is the final transformer block.

\paragraph{Design and objective.}
We evaluate $36$ runs spanning two regimes, three adaptation targets, two
objectives, and seeds $42,43,44$. Each predictor is an affine map with
$5{,}310{,}720$ parameters. Training pairs come from the adapted model's own
greedy answers, on which both the seed and the adapted model are run. All
runs use the same $3000$ test examples. Reported dispersion is sample standard deviation
across the three seeds.

The configured predictor splits contain $4096$ training and $256$ validation
examples. These selections and the test share no IDs or exact texts with
one another or with the adaptation optimizer pools. The state-only objective is
\begin{equation}
  \mathcal L_{\mathrm{state}}^M
  =\mathbb E_{(x,t)}\!\left[
    \frac{1}{d}\left\|f_\phi^M(h_{L,t}^0)
    -\delta_{L,t}^M(x)\right\|_2^2\right].
  \label{eq:predictor-mse}
\end{equation}
Each active objective term is normalized by its value for the training
mean-shift baseline. With
\begin{equation}
  \mathcal L_{\mathrm{KL}}^M
  =\mathbb E_{(x,t)}\,
    \mathrm{KL}\!\left(
      p(h^M_{L,t})\,\middle\|\,
      p\bigl(h^0_{L,t}+f_\phi^M(h^0_{L,t})\bigr)
    \right),
\end{equation}
where $p(h)$ applies the frozen final normalization and output head,
including the model's logit soft-capping, followed by a softmax, the combined
objective is
\begin{equation}
  \mathcal L_{\mathrm{MSE+KL}}
  =\frac{\mathcal L_{\mathrm{state}}^M}{s_{\mathrm{state}}}
  +\frac{\mathcal L_{\mathrm{KL}}^M}{s_{\mathrm{KL}}}.
  \label{eq:predictor-mse-kl}
\end{equation}
Neither objective uses SAE features.

The MSE-only runs select the last checkpoint at step $800$. The reported
values therefore characterize the training budget used here, without
establishing convergence.

\paragraph{Metric.}
We report empty-aware set samples-$F_1$, assigning zero to invalid outputs.
The parser accepts comma-separated allowed labels, ignores their order and
removes duplicates. It accepts \texttt{NONE} for an empty prediction.
A valid empty prediction against an empty gold set receives one.
Like the set parser of \Cref{tab:quality}, this metric ignores label order.
Unlike that parser, it scores an unparseable answer zero instead of treating
it as an empty prediction. The continuous adapters here use $m=20$ virtual
tokens, whereas the continuous entries of \Cref{tab:quality} average over
$m\in\{100,200,500\}$.
% The audited metric field is strict_samples_f1. The separate samples_f1
% field treats invalid answers as empty sets and is not reported here.
% Manual 0.617 and native means/SD were supplied by the experiment owner.
% Their answer records are absent, so those scores were not independently
% recomputed in the artifact audit.

\begin{table}[t]
  \centering
  \caption{Predicted final-state corrections at block 25 during free
  generation on $3000$ test examples. Entries are mean and sample SD over
  adaptation seeds $42,43,44$. Native denotes the adapted model. This predictor
  suite reports a separately evaluated seed baseline of $0.617$; the performance
  sweep in \Cref{tab:quality} reports $0.6092$, so we report raw scores rather
  than combine the two baselines. Target adapters use $n=1000$, with $m=20$
  virtual tokens for the continuous methods. The score is empty-aware set
  samples-$F_1$, with invalid outputs assigned zero.}
  \label{tab:shift-predictor}
  \small
  \setlength{\tabcolsep}{4pt}
  \setlength{\tabcolsep}{4pt}
  \begin{tabular}{lccccc}
    \toprule
    & & \multicolumn{2}{c}{Repredict} & \multicolumn{2}{c}{Fixed} \\
    \cmidrule(lr){3-4}\cmidrule(lr){5-6}
    Target & Native & MSE & MSE+KL & MSE & MSE+KL \\
    \midrule
    GEPA   & $0.678\pm.007$ & $0.683\pm.015$ & $0.681\pm.012$ & $0.638\pm.008$ & $0.683\pm.011$ \\
    Prompt & $0.777\pm.013$ & $0.746\pm.001$ & $0.759\pm.006$ & $0.637\pm.015$ & $0.756\pm.003$ \\
    Prefix & $0.805\pm.007$ & $0.744\pm.004$ & $0.765\pm.006$ & $0.506\pm.121$ & $0.755\pm.007$ \\
    \bottomrule
  \end{tabular}
\end{table}

\paragraph{Constant-shift control.}
For an affine predictor $Wh+b$, initialization at $W=0$ and the train mean
shift does not establish that the learned gain comes from $Wh$, since $b$
can also change. We therefore train a bias-only predictor at block 25 with
the MSE+KL objective and evaluate it in the repredict regime on the same
$3000$ test examples. It reaches $0.633\pm.005$, $0.604\pm.006$ and
$0.543\pm.057$ for GEPA, prompt and prefix, against $0.681\pm.012$,
$0.759\pm.006$ and $0.765\pm.006$ for the affine map. Its invalid-answer
rates are $1.6\%$, $6.0\%$ and $18.4\%$. For prompt and prefix tuning the
constant shift stays below the seed score of $0.617$, so their recovered
gains come from the input-dependent term. The low-rank sweep in Appendix~\ref{sec:predictor-rank}
tests the dimension of the input-dependent part.

\subsection{Fixed Regime and Predictor Rank}
\label{sec:predictor-rank}

Fixed and repredict predictors are trained separately and have different
learned matrices. At inference, a fixed predictor computes the shift once
from the seed state at the generation anchor and adds it at every generated
position. Scores therefore compare the resulting trained systems, rather
than two inference modes of one fitted predictor. With MSE, the fixed regime
leaves $1.9\%$, $6.8\%$ and $30.9\%$ of GEPA, prompt and prefix answers
invalid, and the prefix score varies strongly across seeds. With MSE+KL
these rates are $0.6\%$, $0.1\%$ and $0.1\%$
(\Cref{tab:shift-predictor}).

The low-rank predictor is $f_\phi^M(h)=U_MV_M^\top h+b_M$ with
$U_M,V_M\in\mathbb R^{d\times r}$, initialized like the full predictor with
the train mean shift as bias. \Cref{tab:predictor-rank} reports one training
seed per cell at the final block, trained with MSE and evaluated in both
regimes. The full-rank column repeats the seed-42 run of the same configuration.
These values describe a single seed and carry no uncertainty estimate. Rank
one restricts the input-dependent term to a single direction, giving $0.634$,
$0.596$ and $0.677$ for GEPA, prompt and prefix. This resembles the
single-direction design of prompt steering replacement
models~\citep{heyman2026steer}, but our affine correction includes a vector
bias and does not use their ReLU gate. It is a rank ablation of our
predictor, not a reproduction of PSR.

\begin{table}[ht]
  \centering
  \caption{Task score by predictor rank, seed 42, MSE objective, $3000$ test
  examples. The metric matches \Cref{tab:shift-predictor}, and the manual
  baseline is $0.617$. All predictors operate at block 25.}
  \label{tab:predictor-rank}
  \small
  \begin{tabular}{llccccc}
    \toprule
    Regime & Target & $r=1$ & $r=2$ & $r=4$ & $r=8$ & full \\
    \midrule
    Repredict & GEPA   & 0.634 & 0.640 & 0.645 & 0.660 & 0.683 \\
              & Prompt & 0.596 & 0.655 & 0.664 & 0.699 & 0.746 \\
              & Prefix & 0.677 & 0.683 & 0.688 & 0.700 & 0.741 \\
    \midrule
    Fixed     & GEPA   & 0.635 & 0.635 & 0.636 & 0.640 & 0.641 \\
              & Prompt & 0.580 & 0.577 & 0.595 & 0.608 & 0.619 \\
              & Prefix & 0.353 & 0.346 & 0.371 & 0.391 & 0.389 \\
    \bottomrule
  \end{tabular}
\end{table}

Under repredict, the score increases with rank for every target, while rank 8
remains between $0.023$ and $0.047$ below the full map. Under the fixed
regime the spread across ranks is at most $0.045$, and the fixed prefix
predictor stays far below the seed baseline at every rank.

\subsection{Early Exit}
\label{sec:early-exit}

The early-exit model runs only blocks $0$ to $20$ of the seed-context model.
After block 20 it forms
\begin{equation}
  \widehat h_{25,t}^M=h_{20,t}^0+f_\phi^M(h_{20,t}^0)
\end{equation}
and applies the frozen final normalization and output head. Blocks 21 to 25
are not executed. The map $f_\phi^M$ is a full-rank affine map with
$5{,}310{,}720$ parameters, trained with the MSE+KL objective of
\Cref{eq:predictor-mse-kl} with the adapted block-25 state as target. It is
applied at the prompt and at every generated position, as in the repredict
regime. \Cref{tab:early-exit} compares it with the block-25 corrections of
\Cref{tab:shift-predictor} on the same $3000$ test examples. Fitting the
same map to the seed model's own block-25 state gives $0.618$, against
$0.619$ for the full seed model in the same evaluation (seed 42). For this
task the last five blocks can thus be replaced by an affine map even without
adaptation.

\begin{table}[ht]
  \centering
  \caption{Early exit from block 20 against the block-25 correction, $3000$
  test examples, mean and sample SD over seeds $42,43,44$. The metric matches
  \Cref{tab:shift-predictor}.}
  \label{tab:early-exit}
  \small
  \begin{tabular}{lccc}
    \toprule
    Target & Native & Block-25 correction & Early exit \\
    \midrule
    GEPA   & $0.678\pm.007$ & $0.681\pm.012$ & $0.688\pm.011$ \\
    Prompt & $0.777\pm.013$ & $0.759\pm.006$ & $0.766\pm.012$ \\
    Prefix & $0.805\pm.007$ & $0.765\pm.006$ & $0.770\pm.012$ \\
    \bottomrule
  \end{tabular}
\end{table}

\paragraph{Latency.}
We time generation on the first $100$ test examples,
batch size $16$, one warm-up batch and two timed repeats, excluding model
loading. Every condition decodes exactly $16$ tokens, so the comparison does
not depend on answer length. \Cref{tab:early-exit-latency} reports the
mean time per example. Early exit takes about $43$~ms for all three targets,
against $52.6$~ms for the full seed model, a $1.22\times$ speedup close to
the share of skipped blocks. The speedup over native prompt and prefix tuning
is $1.28\times$ and $1.26\times$. Native GEPA is $7.1\times$ slower than early
exit, mostly because its instruction has $1684$ prompt tokens per example
against $158$ for the seed. This timing run is separate from the quality
evaluation and uses a single seed.

\begin{table}[ht]
  \centering
  \caption{Time per example with $16$ generated tokens, $100$ test examples,
  batch size $16$, mean over two repeats.}
  \label{tab:early-exit-latency}
  \small
  \begin{tabular}{lcc}
    \toprule
    Condition & Native (ms) & Early exit (ms) \\
    \midrule
    Seed   & $52.6$  & -- \\
    GEPA   & $303.7$ & $43.0$ \\
    Prompt & $55.3$  & $43.0$ \\
    Prefix & $54.3$  & $43.1$ \\
    \bottomrule
  \end{tabular}
\end{table}

\subsection{Geometry of the Learned Corrections}
\label{sec:predictor-geometry}

We compare the full-rank repredict predictors across methods. For two
predictors we compute the cosine between their flattened matrices $W_M$.
Cross-method entries pair predictors with the same seed and average over
seeds $42,43,44$. Within-method entries average over the three seed pairs
of one method.

\begin{table}[ht]
  \centering
  \caption{Cosine between flattened block-25 predictor matrices $W_M$
  (repredict, full rank).}
  \label{tab:predictor-geometry}
  \small
  \begin{tabular}{lcc}
    \toprule
    Pair & MSE & MSE+KL \\
    \midrule
    GEPA, prompt   & 0.32 & 0.29 \\
    GEPA, prefix   & 0.27 & 0.24 \\
    Prompt, prefix & 0.49 & 0.57 \\
    \midrule
    GEPA seeds     & 0.72 & 0.67 \\
    Prompt seeds   & 0.67 & 0.76 \\
    Prefix seeds   & 0.61 & 0.68 \\
    \bottomrule
  \end{tabular}
\end{table}

The sample-level analysis uses block 25, MSE and seed 42, the configuration
for which trajectory caches are available. The repredict predictions have
mean cosines of $0.935$, $0.942$ and $0.909$ with the true shifts for GEPA,
prompt and prefix tuning. On the first generated token, the median cosine
between true shifts of two methods is $0.18$, $0.20$ and $0.46$ for the pairs
in \Cref{tab:predictor-geometry}, and the predicted shifts reproduce these
values within $0.05$. Later tokens are generated from different histories
under different methods, and the continuous methods produce only three to
four tokens, so the comparison is restricted to the first steps.

The mean direction carries $58.6\%$, $58.9\%$ and $51.5\%$ of the shift
energy for GEPA, prompt and prefix tuning, the fraction $\eta_\ell$ defined
in \Cref{eq:eta}. The mean directions of different methods have cosines of
$0.09$, $0.13$ and $0.38$. In the repredict predictors, the input-dependent
term $W_Mh$ has mean norms of $180$, $264$ and $315$, against bias norms of
$157$, $212$ and $365$. These norm comparisons do not establish that the bias
or input-dependent term is sufficient on its own.

\subsection{Earlier Composite-Loss Experiment}
\label{sec:legacy-predictor}

Before the MSE ablation, we trained predictors with a composite loss that
includes SAE terms and evaluated them on $512$ held-out examples.

\paragraph{Data and predictors.}
For each adapted model and seed, teacher-forced trajectories of the native
answer pair the seed-model state $h^0_{25,t}$ with the adapted state
$h^M_{25,t}$ at the same position. Training uses $4096$ trajectories and
validation $256$. Evaluation uses $512$ held-out examples and three adaptation
seeds. Generation is greedy with at most $64$ new tokens. The seed-model
samples-$F_1$ is $0.624$.

The affine predictor is $f_\phi(h)=Wh+b$, initialized with $W=0$ and $b$
equal to the training mean shift. The second predictor is a GELU MLP with
hidden width $4608$. Their parameter counts are approximately $5.3$M and
$21$M, respectively. Neither is a small linear probe.

\paragraph{Application modes.}
Let $\tau$ be the generation anchor. One-shot predicts the shift from $h^0_{25,\tau}$
and applies it only at the anchor. Fixed predicts from the same state and
reuses that vector at each decoded position. Repredict computes a new shift
from the current state at each step. Both fixed and repredict are
input-dependent. Their scores compare corrections reused across positions
with corrections recomputed at each position. They do not test whether
conditioning on the example is necessary.

\paragraph{Objective.}
The earlier loss is
\begin{equation}
  \mathcal L=\frac{\mathcal L_{\mathrm{DENSE}}}{s_{\mathrm{DENSE}}}
   +\frac{\mathcal L_{\mathrm{ENC}}}{s_{\mathrm{ENC}}}
   +\frac{\mathcal L_{\mathrm{DEC}}}{s_{\mathrm{DEC}}}
   +\frac{\mathcal L_{\mathrm{KL}}}{s_{\mathrm{KL}}},
  \label{eq:predictor-loss}
\end{equation}
where $\mathcal L_{\mathrm{DENSE}}$ is state MSE, and
$\mathcal L_{\mathrm{ENC}}$ and $\mathcal L_{\mathrm{DEC}}$ are the
corresponding errors in the encoder and decoder space of a Gemma Scope
layer-25 SAE with $16$k latents. The KL term compares adapted and patched
output distributions through the frozen final norm, unembedding and logit
soft-capping, with the adapted distribution as its first argument. Each
$s_k$ is the value of that term for the mean-shift baseline on the training
split. All trajectory positions enter the loss. In one-shot training only
the anchor receives a gradient, so objective values are not comparable with
the recurrent regimes. Checkpoints are selected by validation objective.

\begin{table}[t]
  \centering
  \caption{Strict samples-$F_1$ of the seed model patched at block 25 with a
  predicted shift, mean $\pm$ SD over three adaptation seeds on the same $512$
  test examples. Native is the adapted model itself. The seed model scores
  $0.624$.}
  \label{tab:legacy-shift-predictor}
  \small
  \begin{tabular}{llcccc}
    \toprule
    Method & Predictor & Native & One-shot & Fixed & Repredict \\
    \midrule
    \multirow{2}{*}{GEPA}
      & Linear & \multirow{2}{*}{$0.684\pm.004$}
      & $0.641\pm.009$ & $0.694\pm.007$ & $0.690\pm.008$ \\
      & MLP    & & $0.630\pm.009$ & $0.695\pm.018$ & $0.692\pm.009$ \\
    \midrule
    \multirow{2}{*}{Prompt}
      & Linear & \multirow{2}{*}{$0.838\pm.004$}
      & $0.663\pm.004$ & $0.729\pm.012$ & $0.761\pm.010$ \\
      & MLP    & & $0.648\pm.005$ & $0.747\pm.005$ & $0.744\pm.009$ \\
    \midrule
    \multirow{2}{*}{Prefix}
      & Linear & \multirow{2}{*}{$0.851\pm.006$}
      & $0.657\pm.007$ & $0.755\pm.005$ & $0.765\pm.007$ \\
      & MLP    & & $0.637\pm.009$ & $0.757\pm.004$ & $0.754\pm.011$ \\
    \bottomrule
  \end{tabular}
\end{table}

In this earlier experiment, recurrent scores are higher than one-shot scores
(\Cref{tab:legacy-shift-predictor}).
The largest difference between affine and MLP predictors in a recurrent cell
is $0.018$. The largest fixed-versus-repredict difference is $0.032$. These
observed differences do not establish equivalence. The MSE results in
\Cref{tab:shift-predictor} show that transfer is possible without SAE
supervision.

\subsection{Recurrent Mean Shift}
\label{sec:recurrent-mean}

For prefix tuning, we add the dataset-mean shift at every decoded position
on $500$ intervention-validation examples, choosing its layer and scale on
the same examples. With the selected configuration, samples-$F_1$ rises
from $0.606$ to $0.620$, $0.625$ and $0.627$ for seeds $42$, $43$ and $44$,
while the adapted prefix models score $0.818$, $0.800$ and $0.811$. This
recovers $6.7\%$, $10.0\%$ and $10.0\%$ of the prefix gain.

At block 25, adding the raw train mean with scale one lowers task score below
the seed for all six tested adapted models.

\section{GEPA Prompt Examples}
\label{sec:gepa-prompts}

Inspection of the frozen instructions in the GEPA sweep found
that examples from GEPA's validation data were sometimes copied directly
into the prompt. This observation concerns the optimizer's validation
examples, not evidence of overlap with the held-out test set. Below is the
full $n=200$ instruction reproduced in the workshop draft, with identity
terms in the quoted comments masked by asterisks. The quoted instruction is
preserved verbatim.

\paragraph{GEPA with $n=200$.}
\begin{quote}\small
Classify the supplied text for abusive or harmful language. Determine every label supported by the text itself, using context and the actual target. Do not infer intent or add labels merely because a statement is political, controversial, negative, or mentions a protected group.

Apply labels independently:

\begin{itemize}
\item toxicity: Use for hostile, contemptuous, demeaning, strongly antagonistic, or abusive rhetoric. This is the broad label for toxic language and often co-occurs with insult. Sarcasm, ridicule, dismissive mockery, and hostile political rhetoric can be toxic even without profanity, a threat, or a protected-identity target.
\item obscene: Use only for profanity, vulgarity, sexually explicit language, or clearly obscene wording. Rudeness, political aggression, or a non-obscene mention of sex or identity is not enough.
\item threat: Use only when the text expresses, warns of, or explicitly advocates violence, harm, or other serious retaliation against a person or group. Do not infer a threat from anger, condemnation, predictions of punishment, saying someone deserves to suffer, proposing legal consequences, or hostile language without an actual intention or warning of harm.
\item insult: Use for a direct or clearly implied derogatory characterization of a person, group, audience, institution, or their conduct, including name-calling, belittling comparisons, ridicule, or commands to leave. Words such as “stupid,” “ridiculous,” “childish,” “incompetent,” “over his head,” “do nothing,” or similar contemptuous descriptions can qualify when directed at a person, group, argument, or institution. A hostile political characterization can therefore receive both toxicity and insult. Mere disagreement, factual criticism, or a negative policy assessment without belittling language is not automatically an insult.
\item identity\_attack: Use only when hostility, degradation, exclusion, dehumanization, or denial of equal acceptance/treatment is directed at a protected identity or explicitly because of that identity (for example race, ethnicity, nationality, religion, sex, gender identity, or sexual orientation). Mentioning an identity, discussing discrimination, or criticizing beliefs, conduct, institutions, or policies is not by itself an identity attack. Calling someone “******,” “****-*******,” or similar is generally an accusation about conduct or beliefs, not an attack on the person’s protected identity. Political affiliation and ideological disagreement are not protected-identity attacks.
\end{itemize}

Decision guidance:
\begin{enumerate}
\item Judge the wording and target, not whether the underlying political or factual claim is correct.
\item Broad or indirect hostile rhetoric can still be toxic. Political mockery aimed at a politician, party, legislature, or audience is not exempt.
\item Use insult when the wording belittles or ridicules a discernible target. An insult may target an argument or institution as well as an individual.
\item Toxicity and insult commonly apply together when contemptuous language is used. Toxicity may apply without insult when the hostility is general or lacks a clear derogatory characterization.
\item Do not label threat unless there is an actual threat, explicit call for harm, or expressed intention/warning of serious retaliation.
\item Do not label identity\_attack merely because protected-category terms appear, or because the text accuses someone of prejudice.
\item Do not overread mild sarcasm, institutional criticism, or a hypothetical policy proposal as abuse when it contains no clearly degrading or hostile characterization of a target.
\end{enumerate}

Calibration:
\begin{itemize}
\item “Stupid publicity stunt by another do nothing legislature.” → toxicity and insult: “stupid” and “do nothing” are contemptuous characterizations of the legislature.
\item “The idea ... is ridiculous. Anyone who uses this fallacious argument makes themselves look stupid.” → toxicity and insult: the argument and its users are belittled.
\item “I would have thought that a new Brian Mulroney wing at a federal penitentiary (for wayward politicians) ... would be more appropriate. I’m certain that it would fill up fast.” → no labels: sarcastic political commentary/hypothetical institutional criticism, without sufficiently direct abusive language under this task’s calibration.
\item “There is no scientific evidence ... That is to be expected since there is none.” → no labels: criticism or assertion without abusive language.
\item “Trump is in way over his head ... his childish bluster.” → toxicity and insult: direct political mockery and belittling characterization.
\item “As long as you’re a ****** ... He deserves to suffer and no jury would deny him that.” → toxicity only unless other wording independently supplies a label: “******” is an accusation, and wishing or predicting suffering/legal punishment is not by itself a threat.
\item “BDS ... ****-******* and ****** ... crawl back in your cave.” → toxicity and insult; the accusations alone are not identity\_attack, while “crawl back in your cave” is an insulting command.
\end{itemize}

Return only the labels selected by the fixed output contract; if none apply, return NONE. Never provide an explanation.
\end{quote}

% ===== End ICLR/dense_appendix_1 =====

% ===== Inlined from ICLR/moe_appendix =====
% ---------------------------------------------------------------------------
% MoE appendix, input after \clearpage
% ---------------------------------------------------------------------------

\section{Mixture of experts}
\label{app:moe}

\subsection{Setup in full}
\label{app:moe-setup}

\paragraph{Data and metric.}
Civil Comments~\citep{borkan2019nuanced}, formulated as a multi-label choice
among toxicity, obscene, threat, insult and identity attack, scored by set-level
$F_1$ averaged over rows. Two samples of 2000 test rows are used, which share 1993
rows: the frequency-pruning tables use one, in which 660 rows carry no label, and
the super-expert, router and route-replay evaluations the other, with 656. A model
that answers nothing scores $0.330$ on the first and $0.328$ on the second. These
floors are quoted throughout, and any cell sitting on one is a cell whose answers
must be read before its score is used.

\paragraph{Routing maps.}
Logged on 500 validation rows, disjoint from the test set, so that the mask
built from a map is never selected on the data it is evaluated on. Counts are
kept per layer, per expert and per stage of the sequence: chat template and
instruction, comment, virtual tokens, reasoning and answer.
\Cref{fig:moe-displacement} and the displacement numbers of
\Cref{sec:moe-distribution} use the comment stage, the only one every arm
shares.

\paragraph{Load statistics.}
For layer $\ell$, $u_{\ell,e}$ ($\tilde u_{\ell,e}$ for an adapted arm) is the
share of the layer's top-$k$ assignments that expert $e$ receives, summed over
the 500 calibration rows and over prefill and decoding. The Gini coefficient is
taken over all $n_E$ shares of the layer, experts with no assignment included.
The busiest expert's share is $\max_e u_{\ell,e}$, and the routing
displacement of an adapted arm in layer $\ell$ is given in
\Cref{eq:moe-tv}. A dead expert is one that
receives no assignment in the layer over all 500 rows. \Cref{tab:moe-load-full}
reports the median over layers for the Gini and the busiest share and the mean
over layers for the number of dead experts.

\paragraph{Depth profile of the displacement.}
\label{app:moe-depth}
Averaged over the first and the last quarter of the layers, $\mathrm{TV}_\ell$
falls from $0.11$--$0.17$ to $0.05$--$0.08$ for both continuous methods on Qwen
and from $0.09$--$0.11$ to $0.06$ for prompt tuning on gpt-oss, at all three
budgets. GEPA stays between $0.03$ and $0.04$ at every depth on Qwen, but on
gpt-oss it grows from $0.03$ in the first quarter to $0.07$ in the last, slightly
above prompt tuning there. The gpt-oss
prefix rises from $0.10$--$0.15$ to $0.23$--$0.28$. The odd layers of gpt-oss
use dense attention and the even ones a 128-token band; the prefix displaces the
dense layers by $0.18$--$0.19$ on average against $0.12$--$0.13$ for the banded
ones, whereas prompt tuning ($0.09$--$0.10$ against $0.08$--$0.10$) and GEPA
($0.05$ for both) do not separate them. In the last quarter, the prefix
displacement $\tilde u_{\ell}-u_{\ell}$ has cosine $0.45$--$0.61$ with the base's
shift from comment to reasoning tokens, $u^{\mathrm{reasoning}}_{\ell}-u_{\ell}$,
and $0.35$--$0.44$ with its shift to answer tokens, both taken from the
stage-resolved maps. For prompt tuning and GEPA these cosines lie between
$-0.16$ and $0.16$, so only the prefix moves the late layers toward the experts
the base uses for generated text.

\paragraph{Decoding.}
Greedy on both backbones, no temperature and no nucleus sampling, so a
difference between two arms cannot be sampling noise. The frequency-pruning
runs in \Cref{tab:moe-pruning,tab:moe-pruning-full}, their unpruned level
included, use at most 512 generated tokens on Qwen and 1024 on gpt-oss. The
super-expert-only column uses 64 tokens on Qwen and 1024 on gpt-oss
(\Cref{app:moe-super}). Every gpt-oss arm runs at reasoning effort low; the base
and GEPA reason, and the continuous arms answer directly
(\Cref{sec:moe-imbalance}).

\paragraph{Arms.}
GEPA optimises the instruction text. Prompt tuning learns virtual tokens at the
embedding. Prefix tuning learns per-layer key and value prefixes. The virtual
token budgets are 100, 200 and 500 for both continuous methods, and all
checkpoints share seed 42. Every pruning run checks that its adapter's hash equals
the one recorded in the arm's routing map; for Qwen prefix-500 this is the
checkpoint at step 1125.

\subsection{Complete pruning results}
\label{app:moe-pruning}

\Cref{tab:moe-pruning-full} lists every level of the sweep. The rows marked
``SE kept'' repeat the $-50\%$ mask of the two Qwen arms that collapse there,
with the super experts exempted and the same number of other experts removed
instead; the removed routing mass is approximately unchanged.
The masks are cut from maps rendered at the map harness's default reasoning
effort, medium, while every score uses effort low. The gpt-oss GEPA $-75\%$ cell
sits on this choice: under the medium-effort map, expert 5 of layer 6 is the 26th
least used of its layer and is kept, and $F_1$ is $0.422$ with 36 percent of the
answers readable. A mask cut from a low-effort map of the same arm differs
in 14 of 576 experts and removes this expert as the 24th least used.

\begin{table}[h]
  \centering
  \footnotesize
  \setlength{\tabcolsep}{3.5pt}
  \caption{Frequency pruning, all levels. Mass is the routing mass removed,
  averaged over layers. Empty (answers with no label, unreadable ones included),
  unreadable and no end marker are shares of the 2000 test answers, looping (a
  phrase repeated until the budget runs out) of the first 300. Pruned runs use 512
  generated tokens on Qwen and 1024 on gpt-oss (reasoning effort low on every arm).}
  \label{tab:moe-pruning-full}
  \begin{tabular}{lllccccccc}
    \toprule
    model & arm & level (\%) & mass (\%) & $F_1$ & exact & empty & unreadable & no end & looping \\
    \midrule
    \multirow{19}{*}{Qwen}
      & base         & $0$             &  0.0 & 0.466 & 0.211 &  26.5 &   0.0 &   0.0 &   0.0 \\
      & base         & $-50$           &  5.8 & 0.575 & 0.167 &  21.1 &   0.0 &   0.1 &   0.0 \\
      & base         & $-75$           & 28.3 & 0.330 & 0.330 & 100.0 & 100.0 &  61.9 &  21.7 \\
      & GEPA         & $-50$           &  2.0 & 0.728 & 0.583 &  24.1 &   0.0 &   0.0 &   0.0 \\
      & GEPA         & $-75$           & 17.8 & 0.330 & 0.330 & 100.0 & 100.0 & 100.0 &  94.7 \\
      & prompt, 100  & $-50$           & 11.4 & 0.814 & 0.696 &  31.8 &   0.0 &   0.0 &   0.0 \\
      & prompt, 100  & $-75$           & 35.7 & 0.440 & 0.026 &   3.0 &   0.2 &  29.1 &  28.7 \\
      & prompt, 200  & $-50$           & 10.9 & 0.330 & 0.330 & 100.0 & 100.0 &  90.5 &  59.3 \\
      & prompt, 200  & $-50$, SE kept  & 11.0 & 0.803 & 0.683 &  27.2 &   0.0 &   0.0 &   0.0 \\
      & prompt, 200  & $-75$           & 32.8 & 0.330 & 0.330 & 100.0 & 100.0 &  93.8 &   6.3 \\
      & prompt, 500  & $-50$           & 12.3 & 0.812 & 0.692 &  31.1 &   0.0 &   0.0 &   0.0 \\
      & prompt, 500  & $-75$           & 32.3 & 0.450 & 0.389 &  83.9 &   0.0 &   0.0 &   0.0 \\
      & prefix, 100  & $-50$           &  5.9 & 0.815 & 0.714 &  36.7 &   0.0 &   0.0 &   0.0 \\
      & prefix, 100  & $-75$           & 27.7 & 0.383 & 0.368 &  93.1 &   0.3 &   0.0 &   0.0 \\
      & prefix, 200  & $-50$           &  6.0 & 0.611 & 0.557 &  67.2 &   0.0 &   0.0 &   0.0 \\
      & prefix, 200  & $-75$           & 28.0 & 0.330 & 0.330 & 100.0 & 100.0 & 100.0 &  18.0 \\
      & prefix, 500  & $-50$           &  5.5 & 0.330 & 0.330 & 100.0 & 100.0 &  96.8 &  85.3 \\
      & prefix, 500  & $-50$, SE kept  &  5.5 & 0.768 & 0.654 &  43.0 &   0.0 &   0.0 &   0.0 \\
      & prefix, 500  & $-75$           & 27.7 & 0.330 & 0.330 & 100.0 & 100.0 &  99.8 &  15.0 \\
    \midrule
    \multirow{16}{*}{gpt-oss}
      & base         & $-50$           & 13.3 & 0.565 & 0.422 &  59.4 &  19.6 &   1.3 &   4.7 \\
      & base         & $-75$           & 39.8 & 0.388 & 0.309 &  76.3 &  34.5 &  24.1 &  31.3 \\
      & GEPA         & $-50$           & 10.3 & 0.746 & 0.616 &  36.5 &   0.3 &   0.2 &   3.7 \\
      & GEPA         & $-75$           & 33.7 & 0.422 & 0.374 &  87.4 &  64.4 &  44.2 &  65.0 \\
      & prompt, 100  & $-50$           & 12.2 & 0.502 & 0.338 &  62.9 &  42.4 &  41.5 &  63.7 \\
      & prompt, 100  & $-75$           & 33.6 & 0.330 & 0.325 &  98.3 &  98.2 &  49.6 &  55.3 \\
      & prompt, 200  & $-50$           & 10.4 & 0.584 & 0.449 &  58.5 &  26.2 &  26.2 &  35.3 \\
      & prompt, 200  & $-75$           & 29.7 & 0.343 & 0.326 &  95.7 &  84.3 &  84.1 &  96.7 \\
      & prompt, 500  & $-50$           &  9.1 & 0.607 & 0.401 &  49.0 &  14.3 &  14.2 &  32.0 \\
      & prompt, 500  & $-75$           & 25.5 & 0.334 & 0.323 &  97.2 &  96.8 &  96.7 &  96.0 \\
      & prefix, 100  & $-50$           &  9.5 & 0.736 & 0.655 &  50.9 &   0.0 &   0.0 &   0.0 \\
      & prefix, 100  & $-75$           & 32.6 & 0.528 & 0.450 &  75.7 &   0.9 &   0.0 &   0.0 \\
      & prefix, 200  & $-50$           & 10.3 & 0.793 & 0.677 &  36.6 &   0.0 &   0.0 &   0.0 \\
      & prefix, 200  & $-75$           & 34.6 & 0.486 & 0.332 &  69.0 &   3.0 &   0.0 &   1.0 \\
      & prefix, 500  & $-50$           &  9.8 & 0.797 & 0.673 &  39.0 &   0.0 &   0.0 &   0.0 \\
      & prefix, 500  & $-75$           & 31.7 & 0.405 & 0.381 &  91.0 &   1.1 &   0.0 &   0.0 \\
    \bottomrule
  \end{tabular}
\end{table}

\subsection{Super experts}
\label{app:moe-super}

\paragraph{Identification.}
We use the criterion of \citet{su2026superexperts}: over the first three
quarters of the layers, an expert qualifies when the largest absolute output of
its down-projection exceeds the 99.5th percentile over all experts and a tenth of
the overall maximum, on 32 WikiText-2 segments of 2048 tokens. The statistic has
to be taken inside the fused expert computation of the runtime used here, where a
hook on per-expert modules records nothing. On Qwen3-30B-A3B the criterion returns
the published set, expert 92 of layer 2, 68 of layer 1 and 82 of layer 3 (732, 456
and 428), and on the Instruct-2507 checkpoint the same three (840, 468 and 466).
On gpt-oss-20b it returns expert 5 of layer 17 and expert 5 of layer 6 ($54{,}272$
and $43{,}008$). All reach their maximum on the first position of a raw-text
segment; under the chat template the Qwen experts reach it on position 1, the role
token, in all 232 inputs checked, whatever the content.

\paragraph{Traffic and the frequency masks.}
\Cref{tab:moe-super-removed} lists the super experts each mask removes, read
from the applied masks for eight cells and recomputed from the calibration maps
with the sweep's rule for the other twenty-four; the recomputation reproduces
all eight applied masks exactly. Qwen's sit in the
middle and the tail of their layers and are removed in every arm, expert 68 of
layer 1 already at $-50\%$. On gpt-oss, expert 5 of layer 17 is among the three
busiest experts of its layer in every arm and is never removed, and expert 5 of
layer 6 is removed in four of the sixteen cells, and also in GEPA's $-75\%$
mask when its map is calibrated at reasoning effort low (\Cref{app:moe-pruning}).

\begin{table}[h]
  \centering
  \small
  \setlength{\tabcolsep}{4pt}
  \caption{Super experts removed by the frequency masks of \Cref{tab:moe-pruning}.}
  \label{tab:moe-super-removed}
  \begin{tabular}{lcccc}
    \toprule
    & \multicolumn{2}{c}{Qwen3-30B-A3B} & \multicolumn{2}{c}{gpt-oss-20b} \\
    \cmidrule(lr){2-3}\cmidrule(lr){4-5}
    & $-50\%$ & $-75\%$ & $-50\%$ & $-75\%$ \\
    \midrule
    base         & L1\,E68, L2\,E92 & all three & --- & --- \\
    GEPA         & L1\,E68 & all three & --- & --- \\
    prompt, 100  & L1\,E68 & L1\,E68, L3\,E82 & --- & L6\,E5 \\
    prompt, 200  & L1\,E68, L2\,E92 & L1\,E68, L2\,E92 & --- & --- \\
    prompt, 500  & L1\,E68 & L1\,E68, L2\,E92 & --- & --- \\
    prefix, 100  & L1\,E68 & L1\,E68 & --- & --- \\
    prefix, 200  & L1\,E68, L2\,E92 & all three & L6\,E5 & L6\,E5 \\
    prefix, 500  & all three & all three & --- & L6\,E5 \\
    \bottomrule
  \end{tabular}
\end{table}

\paragraph{Magnitudes and pruning.}
The magnitude of a super expert is the largest absolute output of its
down-projection over 200 Civil test rows. It counts as kept when that value stays
within a factor of ten of the frozen base's. The super experts are pruned alone by
zeroing their down-projection, and the task is scored with at most 64 generated
tokens on Qwen and 1024 on gpt-oss. As in the frequency sweep, an unreadable
answer counts as an empty label set. The Qwen prefix trained at learning rate
$10^{-4}$ is an additional evaluated control, not the selected prefix-500
checkpoint in the frequency sweep; all three of its super-expert magnitudes
are below 1. Seeds 43 and 44 of the same recipe, each at its own
validation-selected checkpoint, keep all three: $440$, $756$ and $456$, and
$185$, $680$ and $452$. Seeds 43 and 44 of prompt tuning at 500 tokens give
$1248$, $780$, $528$ and $584$, $828$, $520$, against $1000$, $952$, $572$ for
seed 42, with their peaks on virtual tokens.

\begin{figure}[h]
  \centering
  \includegraphics[width=0.8\textwidth]{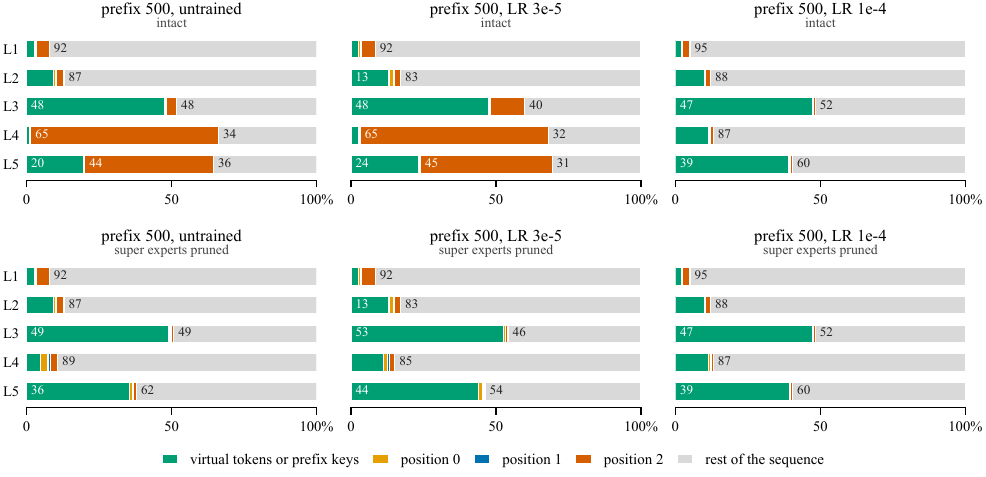}
  \caption{As \Cref{fig:moe-attention}, for a Qwen prefix of 500 tokens, untrained
  and trained at two learning rates.}
  \label{fig:moe-prefix-lr}
\end{figure}

\paragraph{Where the attention goes.}
Attention shares are taken from the eager attention weights, per head and query,
and averaged over heads and over queries from position 3 on, since the first
positions can attend only to themselves, over whole sequences of 200 test rows.
Pruning zeroes the down-projections of the super experts, as in
\Cref{tab:moe-pruning}. gpt-oss returns its weights without the learned sink
column, so the sink's share is the weights' deficit from one; its panels show the
full-attention layers 7 to 17. On Qwen the super experts build the sink from the
second layer on, once expert 68 has written the peak: pruning them leaves the
first layer unchanged and sends nearly all of the sink's attention to the rest of
the sequence (\Cref{fig:moe-attention}). Prompt tuning puts the peak on a virtual
token and concentrates the attention on about twenty of its 500 virtual tokens.
An untrained prefix already moves peak and sink to position 2, and the prefix
trained at $10^{-4}$ has neither, intact or pruned (\Cref{fig:moe-prefix-lr}).

\begin{figure}[h]
  \centering
  \includegraphics[width=\textwidth]{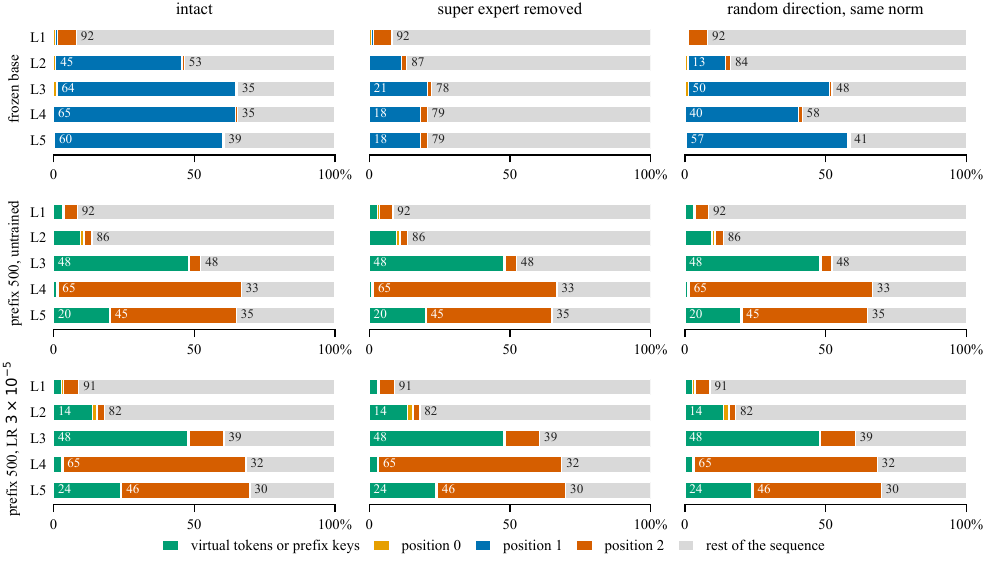}
  \caption{Attention shares on Qwen, layers 1 to 5, 200 Civil test rows: intact,
  with the super expert removed, and with a random perturbation of the same norm.}
  \label{fig:moe-attention-qwen-civil}
\end{figure}

\begin{figure}[h]
  \centering
  \includegraphics[width=\textwidth]{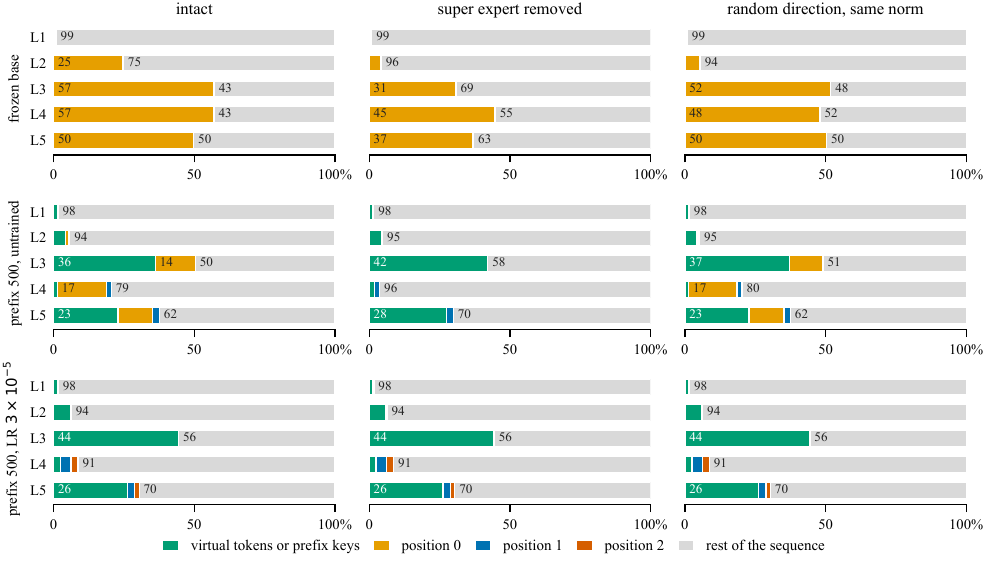}
  \caption{As \Cref{fig:moe-attention-qwen-civil}, on WikiText-2.}
  \label{fig:moe-attention-qwen-wiki}
\end{figure}

Removing only the channel of expert 68's output that carries the peak, on the
first three positions, already takes two thirds of the sink's share from layer 3
on, whereas a random perturbation of the same norm moves much less of the
attention, on the test rows and on WikiText-2
(\Cref{fig:moe-attention-qwen-civil,fig:moe-attention-qwen-wiki}).

\begin{figure}[h]
  \centering
  \includegraphics[width=\textwidth]{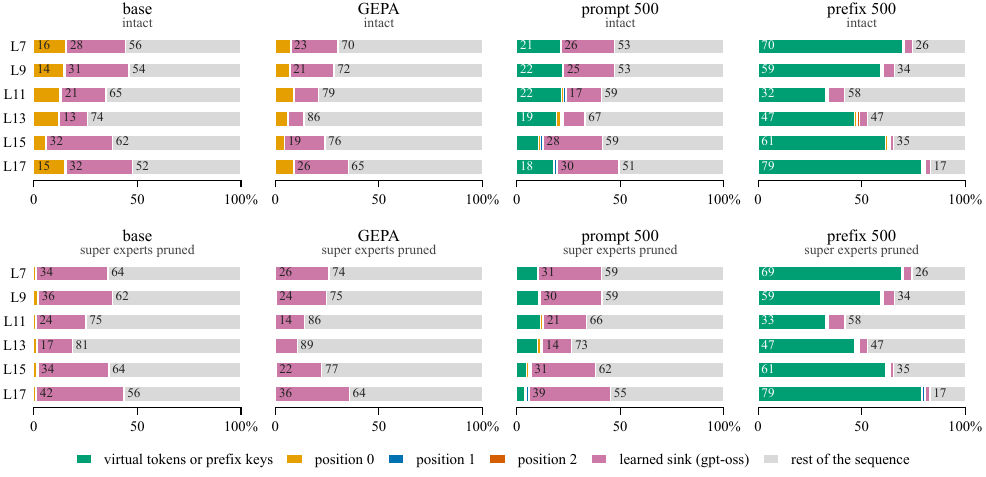}
  \caption{As \Cref{fig:moe-attention}, on gpt-oss, full-attention layers 7 to 17.}
  \label{fig:moe-attention-gptoss}
\end{figure}

On gpt-oss the peak at position 0 is written in layer 6, persists to layer 16 and
collapses at layer 17, the layers of its two super experts. Essentially only its
full-attention layers look at position 0, $12.5\%$ of their attention on average
between layers 7 and 17, while the learned sink takes 13 to 70 percent in every
layer. Pruning the super experts removes the peak, and the attention on position 0
falls to $1.1\%$; of the $11.4$ points it loses, $5.0$ go to the learned sink and
the rest to the sequence (\Cref{fig:moe-attention-gptoss}). This is consistent with the learned sink
supplying the attention that has nowhere else to go~\citep{sun2024massive,gu2025sink},
which on Qwen can only be placed on the peak's token. gpt-oss nonetheless still
builds the peak, which its sink logit, having no value vector, cannot
replace~\citep{queipo2025sinks}; the pruning results say only that this task and
this text do not need it.

\begin{figure}[h]
  \centering
  \includegraphics[width=0.85\textwidth]{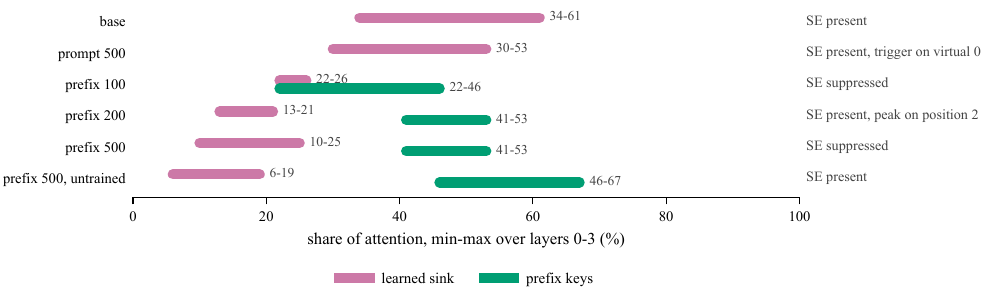}
  \caption{gpt-oss: range over layers 0 to 3 of the attention share taken by the
  learned sink and by prefix keys, per arm, with the state of the super experts.}
  \label{fig:moe-attention-gptoss-arms}
\end{figure}

Across the gpt-oss arms, prefix keys take over part of the learned sink's share in
the first layers, whether the super experts survive or not
(\Cref{fig:moe-attention-gptoss-arms}).

\paragraph{Two prefixes, one score.}
The two Qwen prefixes of 500 tokens that differ only in the learning rate reach the
same $F_1$, but one keeps a super expert and the other none. The one trained at
$10^{-4}$ has neither the peak nor the sink
(\Cref{fig:moe-prefix-lr}) and loses almost nothing when its super experts are
pruned ($0.807$ intact, $0.803$ pruned), but it damages the model most outside the task, through the prefix of its first
six layers (\Cref{tab:moe-prefix-ood}). One method at one score can thus come
with opposite dependence on the super experts. The learning rate does not decide
it alone: two further seeds trained at $10^{-4}$ keep all three super experts
within a factor of three of the base's magnitude (\Cref{app:moe-super}).

\paragraph{Prefix damage outside the task.}
Removing the prefix from the first six layers alone, while it stays in the other
42, brings the WikiText-2 perplexity of every Qwen prefix of \Cref{fig:moe-prefix-lr}
back close to the base's, and the prefix trained at $10^{-4}$ keeps most of its
task gain (\Cref{tab:moe-prefix-ood}). The damage outside the task is carried by
the early layers of the prefix, the task gain mostly by the later ones. The
prefix trained at $3\cdot10^{-5}$ loses its strict score mostly through
unreadable answers, 43 and 32 percent of the 2000 with values zeroed and keys
masked.

\begin{table}[h]
  \centering
  \small
  \setlength{\tabcolsep}{4pt}
  \caption{Qwen prefixes of 500 tokens with the prefix of layers 0 to 5 removed,
  by zeroing its values or masking its keys, at every position. Perplexity on the
  16 WikiText-2 segments of 2048 tokens, strict $F_1$ on the 2000 test rows
  (the untrained prefix was not scored on them). The base scores $6.97$ and
  $0.453$.}
  \label{tab:moe-prefix-ood}
  \begin{tabular}{lcccccc}
    \toprule
    & \multicolumn{3}{c}{WikiText-2 perplexity} & \multicolumn{3}{c}{$F_1$} \\
    \cmidrule(lr){2-4}\cmidrule(lr){5-7}
    prefix & intact & values zeroed & keys masked & intact & values zeroed & keys masked \\
    \midrule
    untrained            & 48.0 & 7.22 & 6.87 & ---   & ---   & ---   \\
    LR $3\cdot10^{-5}$   & 34.0 & 7.48 & 7.16 & 0.807 & 0.412 & 0.509 \\
    LR $10^{-4}$         & 275  & 8.95 & 8.63 & 0.807 & 0.759 & 0.766 \\
    \bottomrule
  \end{tabular}
\end{table}

\subsection{Retraining the router}
\label{app:moe-router}

Only the router of Qwen3-30B-A3B is trained: the 48 gate matrices, $128\times2048$
each and 12.6M parameters in total, with every expert, every adapter and all other
weights frozen. Each arm is trained on its own 1000 training rows and targets, with
cross-entropy on the answer tokens, AdamW at learning rates $10^{-5}$,
$3\cdot10^{-5}$ and $10^{-4}$ (linear warm-up over a tenth of the steps, then
linear decay), batches of eight and four epochs; the optimiser keeps an fp32 copy
of the gates. Epoch 0 is the frozen router. The learning rate and one of the four
trained epochs are chosen on the 200 validation rows on which the arms' own checkpoints were selected, and
the chosen router is scored on the 2000 test rows (\Cref{tab:moe-router}).

\begin{table}[h]
  \centering
  \small
  \setlength{\tabcolsep}{5pt}
  \caption{Retraining only the router on top of each Qwen arm: set-level $F_1$ on 2000
  test rows with the frozen and the retrained router, the paired difference, and the
  routing displacement of \Cref{eq:moe-tv} that the retraining itself causes on the
  comment tokens of the validation rows, with a 64-token generation budget.}
  \label{tab:moe-router}
  \begin{tabular}{lcccc}
    \toprule
    arm & frozen & retrained & difference ($t$) & $\mathrm{TV}$, retrained vs frozen \\
    \midrule
    base        & 0.461 & 0.800 & $+0.339$ ($39.4$) & 0.080 \\
    GEPA        & 0.730 & 0.803 & $+0.073$ ($8.8$) & 0.076 \\
    prompt, 100 & 0.829 & 0.824 & $-0.005$ ($-1.5$) & 0.030 \\
    prompt, 200 & 0.813 & 0.820 & $+0.007$ ($1.6$)  & 0.014 \\
    prompt, 500 & 0.832 & 0.822 & $-0.009$ ($-3.2$) & 0.011 \\
    prefix, 100 & 0.830 & 0.832 & $+0.002$ ($0.3$)  & 0.057 \\
    prefix, 200 & 0.807 & 0.813 & $+0.006$ ($1.4$)  & 0.038 \\
    prefix, 500 & 0.807 & 0.809 & $+0.002$ ($0.2$)  & 0.134 \\
    \bottomrule
  \end{tabular}
\end{table}

On the frozen base, the retrained router recovers most of what adaptation reaches,
while moving less of the comment tokens' routing mass than prompt or prefix tuning
does (\Cref{fig:moe-displacement}). On the continuous arms it adds little: at the
chosen learning rate the test difference lies between $-0.009$ and $+0.007$, at every
learning rate within $\pm0.013$, and a validation gain,
such as $+0.021$ for prefix tuning at 500 tokens, does not carry to the test rows.
GEPA, which routes the comment tokens almost as the base does, gains $0.022$ to
$0.073$ depending on the learning rate, and at the chosen one reaches the $0.80$ of
the retrained base: the router recovers what GEPA gains and GEPA adds nothing on
top of it. Retraining the router and adapting the context are thus substitutes, and
combining them never beats the better of the two by more than $0.007$. This
repeats, on the present checkpoints, what the workshop draft found on another
backbone pair with a different recipe.

\subsection{Expert-parallel timing}
\label{app:moe-ep}

Each of $N$ GPUs holds $E/N$ consecutive experts of every MoE layer and a batch of
tokens. A layer is one real expert-parallel forward: an all-to-all that sends every
token to the devices of its selected experts, a grouped matrix product over the
local experts (SwiGLU, the real expert shapes, random weights), and an all-to-all
back. Its time is the slowest device's, and a step is the sum over layers. Each
token draws its $k$ distinct experts per layer from the arm's measured expert
shares over every routed position from the measured routing maps, so
the per-device load follows the arm's own load. The timing
covers the MoE layers only, without attention or a serving framework, and is not
an end-to-end latency. At 4096 tokens per device, where compute dominates, the
base's own imbalance costs more the fewer experts each device holds, and the
adapted arms stay within 6 percent of the base on two and four devices and within
14 percent on eight (\Cref{tab:moe-ep}); at 512 tokens per device fixed costs
dominate and every adapted arm lies within 6 percent of the base at every device
count.

\begin{table}[h]
  \centering
  \small
  \setlength{\tabcolsep}{4.5pt}
  \caption{Time of the MoE layers per step under expert parallelism relative to the
  base, 4096 tokens per H100 GPU, 20 repetitions; the last row is the base's absolute
  time.}
  \label{tab:moe-ep}
  \begin{tabular}{lcccccc}
    \toprule
    & \multicolumn{3}{c}{gpt-oss-20b} & \multicolumn{3}{c}{Qwen3-30B-A3B} \\
    \cmidrule(lr){2-4}\cmidrule(lr){5-7}
    GPUs & 2 & 4 & 8 & 2 & 4 & 8 \\
    \midrule
    uniform load & 0.956 & 0.853 & 0.707 & 0.975 & 0.938 & 0.821 \\
    GEPA         & 1.010 & 1.040 & 1.109 & 1.008 & 1.046 & 1.062 \\
    prompt, 100  & 1.020 & 1.038 & 1.071 & 1.004 & 1.000 & 0.979 \\
    prompt, 200  & 1.038 & 1.039 & 1.096 & 1.014 & 1.014 & 0.993 \\
    prompt, 500  & 1.042 & 1.061 & 1.135 & 1.019 & 1.026 & 1.007 \\
    prefix, 100  & 1.006 & 1.020 & 1.052 & 1.006 & 1.012 & 1.013 \\
    prefix, 200  & 1.004 & 1.004 & 1.044 & 1.000 & 0.997 & 1.009 \\
    prefix, 500  & 1.006 & 1.029 & 1.062 & 1.003 & 1.004 & 1.011 \\
    \midrule
    base (ms)    & 82.2  & 95.8  & 117.9 & 143.8 & 162.0 & 188.4 \\
    \bottomrule
  \end{tabular}
\end{table}

\subsection{Complete load statistics}
\label{app:moe-load}

The complete load statistics comprise
the expert-by-layer distribution for all sixteen checkpoints, the same
distribution resolved by prompt stage, per-layer Gini with the busiest-expert and
busiest-tenth shares, and, for the pruned runs whose routing was recorded under
the mask, the before-and-after Gini of the surviving experts.
Experts with no traffic are omitted from the distribution tables and counted
separately, since a row of zeros is a fact about the router and not a fact about
the tokens.

\subsection{Replaying routing decisions}
\label{app:moe-carried}

We replay cached routes between the frozen Qwen base and the prompt- and
prefix-tuned arms at 500 virtual tokens, in both directions, on the 2000 test rows.
Each model first generates natively (greedy, at most 64 tokens), and its routes
after top-$k$ selection are cached in every layer. The recipient then generates
again with the donor's routes substituted at the prefill positions of the user
turn, whose token IDs are identical in both models and checked; every other
position, every hidden state and every decoding step remain the recipient's own.
GEPA changes the user turn itself and is not replayed. Three replays separate the
terms of \Cref{eq:moe-layer}: \emph{support} takes the donor's expert set with the
recipient's own router weights renormalised over it, \emph{coefficients} keeps the
recipient's set with weights from the donor's router logits, and \emph{joint}
takes both. A matched random control replaces, per token and layer, exactly as
many of the recipient's experts as the donor's set differs by, with experts drawn
at random, and a shuffled control replays the joint routes of another test row of
the same token geometry (1764 rows have one). Replaying a model's own routes
reproduces its native answers on every row.

On these tokens the adapted arm replaces 21 (prompt) and 22 (prefix) percent of
the base's top-8 experts, changes at least one of them for 87 and 90 percent of
tokens and layers, and the first-ranked one for 37 and 40 percent. The base trails
the arms by $0.373$ and $0.348$ $F_1$ ($0.459$ against $0.832$ and $0.807$ in this
evaluation). Replaying the arm's routes into the base closes none of this gap for
prompt tuning and $0.034$ of it for prefix tuning, where the coefficients alone
give $0.031$ (\Cref{tab:moe-replay}). The expert set alone moves $F_1$ no more than
the matched random control; paired against that control, the joint replay is
higher by $0.011$ $[0.001, 0.021]$ and $0.020$ $[0.009, 0.030]$. Given the base's
routes, the adapted arms lose $0.014$ and $0.023$, against $0.015$ and $0.010$ for
the matched random control. The shuffled routes, which differ from the recipient's
in more of their experts (38 and 39 percent), cost the adapted arms $0.24$ and
$0.39$ and the base $0.06$. Within this scope the rerouting of the user tokens
carries little of the adaptation gain on its own, and undoing it costs about what
any change of its size costs. The replay acts on the prefill of the user turn only
and leaves states and decoding to the recipient, so it does not show that routing
is unnecessary for the adaptation, nor that the remaining state change suffices.

\begin{table}[h]
  \centering
  \scriptsize
  \setlength{\tabcolsep}{3pt}
  \caption{Route replay on Qwen at the prefill of the user turn, 2000 test rows:
  change in the recipient's $F_1$ against its own native answers, with paired
  bootstrap 95\% intervals below. Gap is the donor's native $F_1$ minus the
  recipient's. Shuffled uses the 1764 rows with a donor row of the same geometry.}
  \label{tab:moe-replay}
  \begin{tabular}{lcccccc}
    \toprule
    recipient $\leftarrow$ donor & gap & support & coefficients & joint & matched random & shuffled \\
    \midrule
    base $\leftarrow$ prompt, 500  & $+0.373$ & $-0.014$ & $+0.005$ & $-0.002$ & $-0.012$ & $-0.057$ \\
                                   &          & $[-0.025, -0.003]$ & $[-0.006, +0.015]$ & $[-0.013, +0.010]$ & $[-0.022, -0.002]$ & $[-0.079, -0.035]$ \\
    prompt, 500 $\leftarrow$ base  & $-0.373$ & $-0.011$ & $-0.013$ & $-0.014$ & $-0.015$ & $-0.237$ \\
                                   &          & $[-0.018, -0.004]$ & $[-0.020, -0.005]$ & $[-0.023, -0.005]$ & $[-0.023, -0.007]$ & $[-0.260, -0.215]$ \\
    base $\leftarrow$ prefix, 500  & $+0.348$ & $+0.007$ & $+0.031$ & $+0.034$ & $+0.014$ & $-0.063$ \\
                                   &          & $[-0.002, +0.017]$ & $[+0.022, +0.041]$ & $[+0.022, +0.045]$ & $[+0.005, +0.024]$ & $[-0.085, -0.042]$ \\
    prefix, 500 $\leftarrow$ base  & $-0.348$ & $-0.009$ & $-0.011$ & $-0.023$ & $-0.010$ & $-0.392$ \\
                                   &          & $[-0.019, +0.001]$ & $[-0.021, -0.002]$ & $[-0.034, -0.011]$ & $[-0.020, -0.001]$ & $[-0.419, -0.365]$ \\
    \bottomrule
  \end{tabular}
\end{table}

% ===== End ICLR/moe_appendix =====

\end{document}